\documentclass[sigconf]{acmart}
\usepackage{graphics}
\usepackage{graphicx}
\usepackage{xspace}
\usepackage{subcaption}
\usepackage{amsmath}
\usepackage{amsthm}
\usepackage{mathrsfs}
\usepackage{array}
\usepackage{textcomp}
\usepackage{weiwAlgorithm}
\usepackage{url}
\usepackage{diagbox}
\usepackage{color}
\usepackage{booktabs}
\usepackage{listings}
\usepackage{tcolorbox}
\usepackage{setspace}
\usepackage{multirow}
\usepackage{booktabs}    % \toprule, \midrule, \bottomrule
\usepackage{tabularx}    % tabularx
\usepackage{array}       % column formatting (p{}, m{}, >{\centering} etc.)
\usepackage{makecell}    % line breaks inside table cells
\usepackage{enumitem}

\usepackage{colortbl} % add colours to tables. \rowcolor{}, \columncolor{}, and \cellcolor{}
\usepackage{xcolor} % provides colour definitions and options for LaTeX. It defines colours (like red, blue, gray) and allows mixes like blue!10

\usepackage{longtable}
\usepackage[normalem]{ulem}
\useunder{\uline}{\ul}{}

\newtheorem{definition}{Definition}

\newcommand{\myparagraph}[1]{\vspace{1mm} \noindent \textbf{#1}.}

\newcommand{\myparagraphunderline}[1]{\vspace{0.5mm} \noindent \underline{#1.}}
\newcommand{\myparagraphquestion}[1]{\vspace{1mm} \noindent \textbf{#1?}}

\newcommand{\ie}{{i.e.,}\xspace}

\newcommand{\pokh}{{PRoH}\xspace}
\newcommand{\pokhl}{{PRoH-L}\xspace}
\newcommand{\hgr}{{HyperGraphRAG}\xspace}

\AtBeginDocument{%
  }

\copyrightyear{2026}
\acmYear{2026}
\setcopyright{cc}
\setcctype{by}
\acmConference[WWW '26] {Proceedings of the ACM Web Conference 2026}{April 13--17, 2026}{Dubai, United Arab Emirates}
\acmBooktitle{Proceedings of the ACM Web Conference 2026 (WWW '26), April 13--17, 2026, Dubai, United Arab Emirates}
\acmISBN{979-8-4007-2307-0/2026/04}
\acmDOI{10.1145/3774904.3792611}

\begin{document}

\title{PRoH: Dynamic Planning and Reasoning over Knowledge Hypergraphs for Retrieval-Augmented Generation}

\author{Xiangjun Zai}
\orcid{0009-0009-6997-2127}
\affiliation{
  \institution{University of New South Wales}
  \city{Sydney}
  \country{Australia}}
\email{xiangjun.zai@unsw.edu.au}

\author{Xingyu Tan}
\orcid{0009-0000-7232-7051}
\affiliation{
  \institution{University of New South Wales}
  %\institution{Data61, CSIRO}
  \city{Sydney}
  \country{Australia}}
\email{xingyu.tan@unsw.edu.au}
\authornote{Corresponding author.}

\author{Xiaoyang Wang}
\orcid{0000-0003-3554-3219}
\affiliation{
  \institution{University of New South Wales}
  \city{Sydney}
  \country{Australia}}
\email{xiaoyang.wang1@unsw.edu.au}

\author{Qing Liu}
\orcid{0000-0001-7895-9551}
\affiliation{
  \institution{Data61, CSIRO}
  \city{Hobart}
  \country{Australia}}
\email{q.liu@data61.csiro.au}

\author{Xiwei Xu}
\orcid{0000-0002-2273-1862}
\affiliation{
  \institution{Data61, CSIRO}
  \city{Sydney}
  \country{Australia}}
\email{xiwei.xu@data61.csiro.au}

\author{Wenjie Zhang}
\orcid{0000-0001-6572-2600}
\affiliation{
  \institution{University of New South Wales}
  \city{Sydney}
  %\state{NSW}
  \country{Australia}}
\email{wenjie.zhang@unsw.edu.au}

% \renewcommand{\shortauthors}{Xiangjun Zai et al.}
\renewcommand{\shortauthors}{Xiangjun Zai et al.}
%% No italics, no superscripts, not anonymous
%% Use footnote or author note to identify equal contribution and/or contact author info

\begin{abstract}
Knowledge Hypergraphs (KHs) have recently emerged as a knowledge representation for retrieval-augmented generation (RAG), offering a paradigm to model multi-entity relations into a structured form. However, existing KH-based RAG methods suffer from three major limitations: static retrieval planning, non-adaptive retrieval execution, and superficial use of KH structure and semantics, which constrain their ability to perform effective multi-hop question answering. To overcome these limitations, we propose \textbf{\pokh}, a dynamic \underline{P}lanning and \underline{R}easoning \underline{o}ver Knowledge \underline{H}ypergraphs framework. PRoH incorporates three core innovations: (i) a context-aware planning module that sketches the local KH neighborhood to guide structurally grounded reasoning plan generation; (ii) a structured question decomposition process that organizes subquestions as a dynamically evolving Directed Acyclic Graph (DAG) to enable adaptive, multi-trajectory exploration; and (iii) an Entity-Weighted Overlap (EWO)-guided reasoning path retrieval algorithm that prioritizes semantically coherent hyperedge traversals.
Experiments across multiple domains demonstrate that PRoH achieves state-of-the-art performance, surpassing the prior SOTA model \hgr by an average of {19.73\%} in F1 and {8.41\%} in Generation Evaluation (G-E) score, while maintaining strong robustness in long-range multi-hop reasoning tasks.
  
\end{abstract}




\maketitle

\captionsetup[table]{aboveskip=3pt, belowskip=3pt}

\captionsetup[figure]{aboveskip=3pt, belowskip=3pt}

\vspace{-4mm}
\section{Introduction}
\label{sec:intro}

\begin{figure}[t]
    \centering
    \includegraphics[width=1\linewidth]{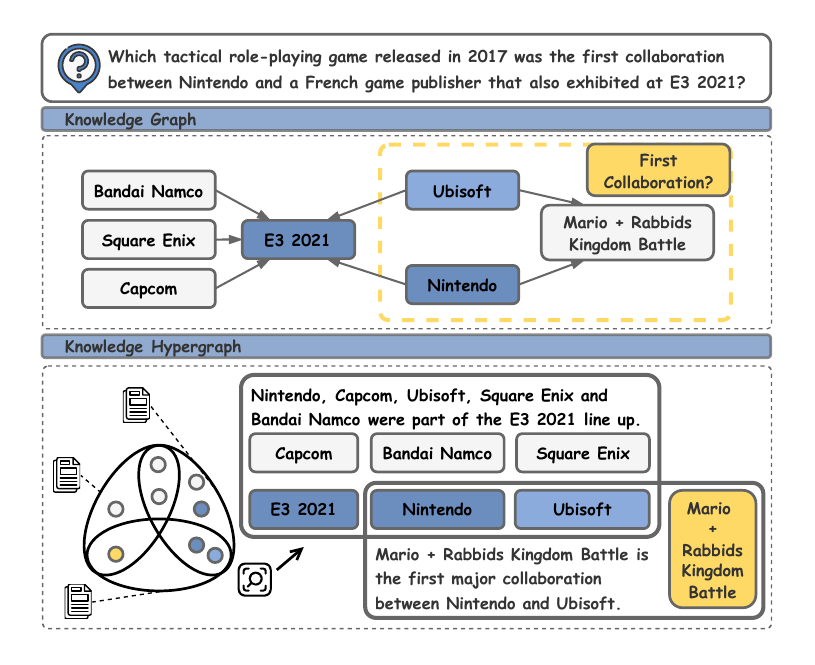}
    \caption{Illustration of Knowledge Hypergraph.}\label{fig:hyperedges}
\end{figure}

To improve the factual accuracy and specificity of large language model (LLM) responses, Retrieval-Augmented Generation (RAG) has emerged as a promising approach that integrates external knowledge through the in-context learning capabilities of LLMs.
However, traditional RAG systems rely primarily on semantic similarity, fail to capture the structured relational knowledge inherent in many information domains, and often retrieve redundant or noisy content~\cite{tan2025hydrarag}.
To address this limitation, Graph-based RAG has been introduced to integrate explicitly structured representations of knowledge into the retrieval process, enabling more accurate and interpretable reasoning \cite{edge2024local,guo2024lightrag,jimenez2024hipporag,jimenez2025hipporag2}. 
By representing entities and their relationships as Knowledge Graphs (KGs), these approaches can capture indirect relations and support multi-hop reasoning across interconnected facts.
Nevertheless, most existing Graph-based RAG frameworks model only relations that involve exactly two entities. 
This design overlooks a fundamental property of real-world knowledge: many relations are inherently n-ary, involving more than two entities simultaneously. 
As shown in Figure~\ref{fig:hyperedges}, the relation "Mario + Rabbids Kingdom Battle is the first major collaboration between Nintendo and Ubisoft." connects three entities: "Mario + Rabbids Kingdom Battle", "Nintendo" and "Ubisoft".
In such cases, the semantics of a relation is established only when all participating entities are considered together. 
Decomposing these n-ary relations into multiple binary edges inevitably causes a loss of critical structural and semantic information \cite{DBLP:conf/ijcai/WenLMCZ16,DBLP:conf/www/GuanJWC19,DBLP:conf/ijcai/FatemiTV020,DBLP:conf/www/RossoYC20}.

To address the representational gap, Knowledge Hypergraphs (KHs) have been proposed as a more compact and semantically expressive knowledge structure for Graph-based RAG~\cite{DBLP:journals/corr/abs-2503-16530,luo2025hypergraphrag,feng2025hyperrag,DBLP:journals/corr/abs-2508-11247}. 
KHs generalize standard KGs by allowing hyperedges to connect multiple nodes simultaneously, naturally capturing complex multi-entity interactions and preserving n-ary relational semantics more faithfully. 
As a result, KHs enhance both the storage of knowledge and its retrieval efficiency for downstream comprehension tasks.
Currently, most KH-based RAG systems follow a three-stage pipeline:
$i)$ KH construction: extract entities and n-ary relations from the text sources to build the KH;
$ii)$ KH-guided retrieval: link query topics with entities and map queries to hyperedges through predefined similarity metrics, followed by a shallow, heuristic-based graph fusion that retrieves graph elements along with their corresponding source text chunks;
$iii)$ Answer generation: pass the retrieved knowledge directly to the generation module.
Although this approach has shown promising results in question-answering (QA) tasks, it does not fully exploit the expressive potential of KHs and suffers from the following limitations:

\myparagraphunderline{Limitation 1: Static Retrieval Planning}
Existing frameworks rely on predefined, hard-coded retrieval pipelines that apply the same sequence of operations regardless of query content or graph context. 
For example, HGRAG~\cite{DBLP:journals/corr/abs-2508-11247} performs a KH diffusion with a fixed number of iterations, without evaluating whether the retrieved passages are sufficient.
This rigid design prevents the model from adapting its retrieval plan to the question semantics or the KH topology, leading to inefficient and misaligned knowledge access.

\myparagraphunderline{Limitation 2: Non-Adaptive Retrieval Execution}
Current systems perform retrieval in a one-shot, non-iterative manner, relying solely on the original query.
For instance, \hgr~\cite{luo2025hypergraphrag} identifies and retrieves relevant entities and hyperedges based on a predefined similarity threshold in one graph call.
Such static execution fails to refine retrieval using intermediate reasoning results, limiting the system's capability for multi-hop reasoning.

\myparagraphunderline{Limitation 3: Superficial Use of Graph Structure and Semantics}
Existing methods primarily treat hyperedges as simple links or routing mechanisms for accessing associated text chunks~\cite{luo2025hypergraphrag, feng2025hyperrag}.
This superficial treatment ignores the rich relational semantics encoded in hyperedges and misses the opportunity to guide more precise retrieval and reasoning within KH-based RAG frameworks.

\myparagraph{Contribution}
To better realize the potential of KHs for RAG, we introduce \textbf{\pokh}, a dynamic KH-based RAG framework that fully leverages the expressive power of KHs.
\pokh performs structured planning and reasoning directly over KHs, enabling the retriever to adaptively explore and integrate knowledge for multi-hop question answering. The key ideas of \pokh can be summarized as follows:

\myparagraphunderline{Context aware planning}
\pokh employs a graph scope-aware planning strategy. Before performing question decomposition, 
\pokh first sketches the local neighborhood of the topic entities within the KH. This pre-planning step provides the LLM with a brief yet informative view of the topological and semantic scope of the question-relevant subgraph, mitigating the mismatch between linguistic-only decomposition and the actual graph knowledge available. Consequently, the LLM produces reasoning plans that are more feasible and better aligned with the structure of the underlying KH.

\myparagraphunderline{Structured question decomposition with iterative refinement}
\pokh adopts a structured question decomposition approach to explicitly capture the dependencies among subquestions.
Instead of treating subquestions as a linear sequence, the reasoning plan is represented as a Directed Acyclic Graph (DAG) that captures logical precedence among them. As reasoning progresses following the topological order of the subquestions, the DAG is iteratively refined. 
To be more specific,
later subquestions are updated, and new subquestions and dependencies may emerge.
This mechanism allows the reasoning plan to evolve dynamically and remain consistent with the current reasoning state.
\pokh also introduces a state-space search mechanism that performs reasoning as a branching exploration from the current reasoning state, effectively modeling the process as a tree. Unlike prior methods that assume each subquestion has one single correct answer, our approach allows multiple candidate answers per subquestion. That is, several reasoning trajectories can coexist. This design corresponds to the multi-entity nature of n-ary relations, allowing \pokh to manage ambiguity and recover from local errors. The state exploration continues until one or more trajectories reach a goal state, where all subquestions are resolved and a final answer can be derived.

\myparagraphunderline{EWO-guided reasoning path retrieval}
\pokh employs a fine-grained reasoning path retrieval strategy guided by the Entity-Weighted Overlap (EWO) score, specifically designed for KHs. When visiting a hyperedge, for each hyperedge that overlaps with the current one, the retriever evaluates how strongly the overlapping entities contribute to answering the current subquestion and aggregates these relevance scores to determine the next traversal direction. 
This process allows the retriever to prioritize semantically meaningful connections rather than relying on purely structural overlaps. 
As a result, the retrieved reasoning paths are better aligned with the underlying semantics of the question, enabling more accurate and interpretable multi-hop reasoning.
In summary, the main contributions of this paper are as follows:
\begin{itemize}[leftmargin=*]
    \item We propose \pokh, a dynamic KH-based RAG framework that fully leverages the expressive power of hypergraphs for multi-hop question answering.
    \item We introduce a context-aware planning mechanism that sketches the underlying KH and generates feasible reasoning plans.
    \item We develop an EWO-guided reasoning path retrieval strategy for fine-grained, semantically aware exploration of KHs.
    \item \pokh consistently achieves better performance and interpretability than the state-of-the-art \hgr framework across multiple knowledge domains, surpassing it by an average of 19.73\% in F1 and 8.41\% in Generation Evaluation (G-E).
\end{itemize}

\begin{figure*}[t]
    \centering
    \includegraphics[width=1\linewidth]{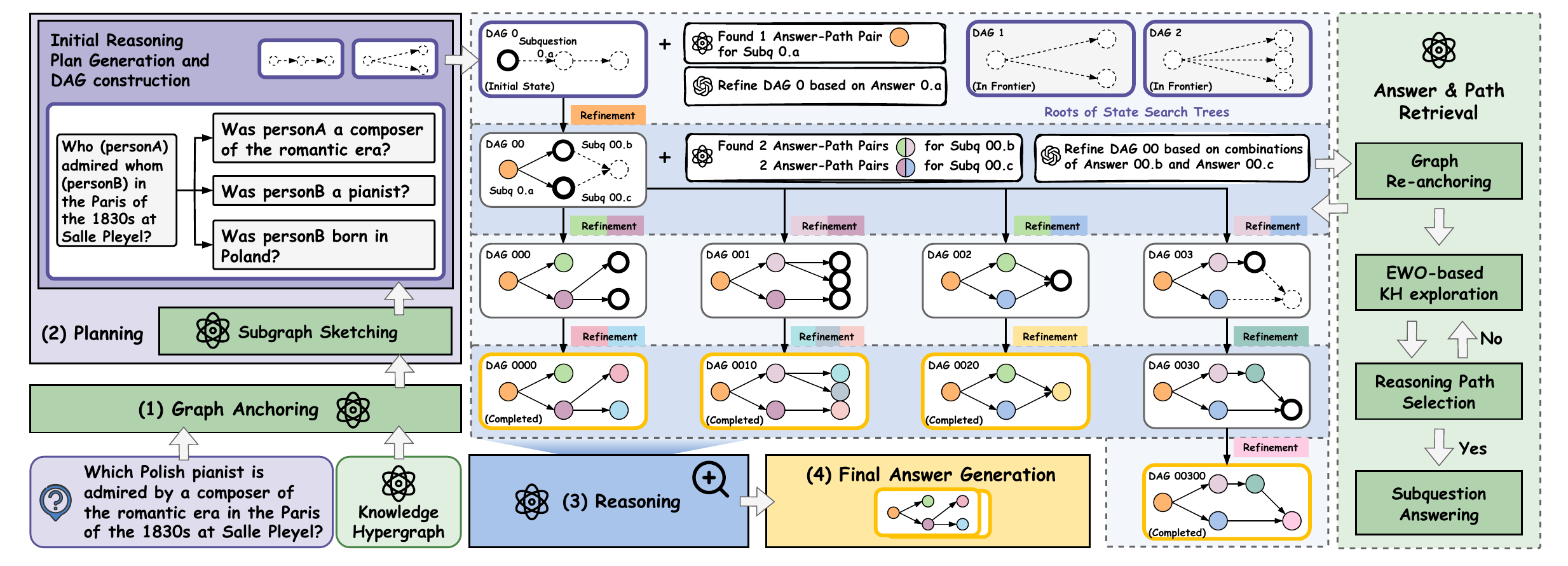}
    \Description{Overview of the \pokh framework.}
    \caption{Overview of the \pokh framework. (1) Given an input question and a KH, \textit{Graph Anchoring} constructs a question subgraph. (2) The following \textit{Planning} stage sketches the question subgraph and obtains context to generate initial reasoning plans and construct reasoning DAGs for the question. (3) The initial reasoning DAGs serve as roots of the state search trees in the \textit{Reasoning} stage. At each state within the tree, the model processes the current reasoning DAG, retrieves answer-path pairs from the KH and iteratively refines the current DAG to transition to the next state until one or more DAGs are completed. (4) The completed DAGs are then passed to the \textit{Final Answer Generation} as the retrieved knowledge for producing the final answer.
    }\label{fig:workflow}
    
\end{figure*}

\section{Related Work}
\label{sec:related_work}

\myparagraph{Graph-based RAG}
Graphs are a natural representation for modeling relational structure among entities~\cite{DBLP:journals/pvldb/WuSWWZQL24, DBLP:conf/icde/WuSWZ00024, DBLP:journals/pacmmod/SimaYWZZL25, tang2025tabular, guo2025efficient, DBLP:conf/icde/WangOWWL25}.
Unlike traditional RAG systems, which rely on flat textual corpora as their primary knowledge source, Graph-based RAG leverages structured representations of relational knowledge.
These systems either construct knowledge graphs from documents or utilize existing large-scale graphs such as Wikidata~\cite{DBLP:journals/cacm/VrandecicK14} and Freebase~\cite{DBLP:conf/sigmod/BollackerEPST08}, enabling more precise retrieval and relation-aware reasoning~\cite{DBLP:journals/corr/abs-2408-08921, huang2025embedding, huang2025survey, wu2023holistic, he2023scaling, he2025robust, wang2023towards, DBLP:journals/tkde/WuSWZQZL25}.
EmbedKGQA~\cite{DBLP:conf/acl/SaxenaTT20} first introduces KG embedding methods into multi-hop KGQA.
Recent representative frameworks~\cite{edge2024local,guo2024lightrag,jimenez2024hipporag,wu2024query2gmm,DBLP:conf/acl/WuZQCXMJG25, zhu-etal-2025-knowledge} extract entities and relations from unstructured text using LLMs and index them for retrieval via graph traversal or embedding-based matching.
More advanced approaches~\cite{DBLP:conf/www/TanWLXYZ25, DBLP:conf/iclr/Sun2024TOG,DBLP:conf/nips/Chen2024plan,DBLP:conf/aaai/Ma0CSWPTSLZC25,DBLP:conf/iclr/MaXJLQYMG25,DBLP:conf/aaai/LiangG25,DBLP:conf/acl/LiuL0W0W0025,DBLP:conf/acl/GaoZALNQL25,ijcai2024p734, DBLP:journals/corr/abs-2502-12029} incorporate iterative reasoning and adaptive planning mechanisms to enhance graph-guided inference. 
Learning-based methods~\cite{DBLP:conf/acl/MavromatisK25,DBLP:conf/iclr/LuoLHP24,10.1609/aaai.v39i24.34747,tan2025memotime,tan2026privgemo} have also shown promising results.
Despite these advances, existing graph-based RAG frameworks remain limited by their binary relational representations \cite{tan2025hydrarag}, which restrict their capacity to model multi-entity facts and capture higher-order relational semantics.

\myparagraph{Knowledge Hypergraph-based RAG}
Hypergraphs are widely adopted to model higher-order relationships~\cite{xie2025hl,zhang2025accelerating,luo2024hierarchical, wang2025llm,wang2025llm2, wang2024simpler,tan2023higher}.
Early systems such as HyperGraphRAG~\cite{luo2025hypergraphrag} and Hyper-RAG~\cite{feng2025hyperrag} extract hyperedges from textual sources to capture higher-order relations and employ similarity-based retrieval to identify relevant entities, hyperedges, and text chunks for question answering.
IdepRAG~\cite{DBLP:journals/corr/abs-2503-16530} leverages the hypergraph structure by performing Personalized PageRank over topic entities to locate contextually relevant evidence.
Meanwhile, HGRAG~\cite{DBLP:journals/corr/abs-2508-11247} introduces a cross-granularity retrieval mechanism that integrates fine-grained entity similarity with coarse-grained passage similarity through hypergraph diffusion, effectively balancing structural precision and semantic coherence.
However, current hypergraph-based approaches still rely on heuristic, one-shot retrieval pipelines and lack context-aware and iterative reasoning capabilities, motivating the framework proposed in this study.

\section{Preliminaries}
\label{sec:preliminary}

Let $\mathcal{H}=(\mathcal{V}, \mathcal{E})$ denote a knowledge hypergraph, where $\mathcal{V}$ and $\mathcal{E}$ represent the set of entities and hyperedges, respectively. 
Each hyperedge $e \in \mathcal{E}$ links a set of entities $V(e) \subseteq \mathcal{V}$, \ie $V(e) = \{v_1, v_2 \cdots, v_n\}$ where $n \geq 1$.
An n-ary relationship fact is modeled as $f = (e, V(e))$.
We denote the set of hyperedges in $\mathcal{E}$ that contains an entity $v$ as $E(v)$, \ie $E(v) = \{e | v \in V(e) \land e \in \mathcal{E}\}$.
A subgraph $\mathcal{H_S}=(\mathcal{V_S},\mathcal{E_S})$ is an induced subgraph of $\mathcal{H}$, if $\mathcal{E_S} \subseteq \mathcal{E}$, $\mathcal{V_S}=\{v| v \in V(e) \land e \in \mathcal{E_S}\}$.
Two hyperedges $e_i$ and $e_j$ are connected iff $V(e_i) \cap V(e_j) \neq \emptyset$, \ie an overlap of entities exists between $e_i$ and $e_j$.
The set of hyperedges connected to hyperedge $e$ is defined as the neighbor edge set $\mathrm{Nbr}(e) = \{e' \mid V(e') \cap V(e) \neq \emptyset \land e' \neq e \land e' \in \mathcal{E}\}$.

\begin{definition}[Hypergraph-based RAG] 
Given a question $q$ and a knowledge hypergraph $\mathcal{H}=(\mathcal{V}, \mathcal{E})$, hypergraph-based RAG retrieves question-related knowledge, \ie a set of facts $\mathcal{F}$ from $\mathcal{H}$ and then generates an answer $a(q)$ based on $q$ and $\mathcal{F}$.
\end{definition}

\begin{definition}[Reasoning Path] 
Given a knowledge hypergraph $\mathcal{H}=(\mathcal{V}, \mathcal{E})$,
a reasoning path within $\mathcal{H}$ is a connected sequence of hyperedges, represented as $path(e_s, e_t) = [e_1=e_s, e_2, \cdots, e_{l-1}, e_l = e_t]$, where $l$ denotes the length of the path, i.e., $l = \mid path(e_s, e_t) \mid$.
\end{definition}

\section{Method}
\label{sec:method}
In this section, we introduce the framework of \pokh.
As illustrated in Figure \ref{fig:workflow}, PRoH operates in four stages: Graph anchoring, Planning, Reasoning with Graph retrieval, and Final answer generation.
It generates and dynamically refines reasoning plans that consist of subquestions decomposed from the original question, and retrieves knowledge from the KHs via exploring reasoning paths.

\subsection{Graph Construction and Indexing} \label{subsec:indexing}
\myparagraph{KH Construction}
We adopt the graph construction method introduced in \hgr~\cite{luo2025hypergraphrag}. Given the documents, the method first extracts hyperedges from text chunks, then identifies entities within these hyperedges, and finally constructs the KH $\mathcal{H}=(\mathcal{V}, \mathcal{E})$.
Each entity is represented as $v = (v^{nm}, v^{desc})$, where $v^{nm}$ is the unique entity name and $v^{desc}$ denotes its associated textual description.
Each hyperedge is represented as $e = (e^{nm}, V(e), e^{ref})$, where $e^{nm}$ is the unique hyperedge name, $V(e) \subseteq \mathcal{V}$ is the set of entities linked by $e$, and $e^{ref}$ is the reference to the originating text chunks.
For efficient retrieval, vector databases are maintained for entity names, entity descriptions, and hyperedge names.

\myparagraph{Synonym Hyperedge Augmentation}
In the original method~\cite{luo2025hypergraphrag}, entity de-duplication relies on exact name matching, which results in isolated hyperedges and weakens the connectivity of the constructed KH. 
To better utilize graph structure in the later retrieval stage, inspired by HippoRAG2~\cite{jimenez2025hipporag2}, we introduce synonym hyperedges to the constructed KH. 
To be more specific, the synonym hyperedges are augmented in three steps.
$i)$ For each pair of entities $(v_i, v_j) \in \mathcal{V} \times \mathcal{V}$, we compute the cosine similarity

\begin{equation}
\mathrm{sim}(v_i, v_j) = \cos \left( \mathbf{z}(v_i^{\text{nm}}), \mathbf{z}(v_j^{\text{nm}}) \right),
\label{eq:cossim}
\end{equation}
\noindent
where $\mathbf{z}(\cdot)$ is the embedding function. 
We add a similarity edge $e_{sim} = (v_i, v_j)$ if $\mathrm{sim}(v_i, v_j) \geq \tau$.
$ii)$ We form the similarity subgraph $\mathcal{H}_{sim}=(\mathcal{V}_{sim}, \mathcal{E}_{sim})$
and compute its connected components $\{C_1, C_2, \dots, C_m\}$, where each $C_i \subseteq \mathcal{V}_{sim}$.
$iii)$ For each connected component $C_i$, we query an LLM with the entity names and descriptions to determine whether they represent synonymous entities. 
If the set of synonymous entities $V_{syn} \subseteq C_i$ is confirmed, we add a synonym hyperedge $e_{\text{syn}}$ which links all entities in $V_{syn}$.

\subsection{Graph Anchoring}\label{subsec:graph_achor}

\myparagraph{Topic Entity Identification}
Given a question $q$, we first utilize an LLM to extract a set of keywords $T_w$. 
Each keyword is linked to a topic entity by computing cosine similarity (Eq.~\eqref{eq:cossim}) against all entities in $\mathcal{V}$.
The highest-scoring entity whose similarity exceeds a threshold $\theta_v$ is selected as the topic entity for the keyword, and all selected entities form the topic entity set $\mathcal{T}$.

\myparagraph{Target Hyperedge Matching}
To exploit the semantic information contained in the question $q$, we further match related hyperedges from $\mathcal{E}$. 
For each hyperedge, we compute a similarity score between $q$ and the hyperedge name, following a formulation analogous to the cosine similarity in Eq.~\eqref{eq:cossim}. The top-ranked hyperedges that satisfy the threshold $\theta_{e}$ form the target hyperedge set $\mathcal{R}$.

\myparagraph{Question Subgraph Construction}
Once the topic entities $\mathcal{T}$ and target hyperedges $\mathcal{R}$ are identified, we construct a question subgraph to constrain subsequent retrieval during planning and reasoning. Specifically, for each $v \in \mathcal{T}$ and $e \in \mathcal{R}$, we extract its $D_{\max}$-hop neighborhood from $\mathcal{H} = (\mathcal{V}, \mathcal{E})$. The question subgraph $\mathcal{H}_q$ is defined as the union of these neighborhoods. We also merge synonymous entities during this subgraph construction phase to obtain a compact representation of the original KH, which benefits the subsequent planning and reasoning.

\subsection{Planning}\label{subsec:planinit}
For multi-hop questions, directly retrieving graph elements from the immediate neighborhood of topic entities or target hyperedges is often insufficient. 
However, naively expanding to deeper neighborhoods quickly leads to an information explosion, as the number of reachable entities and hyperedges grows exponentially with depth. 
This effect is particularly critical in hypergraphs, where n-ary relations link multiple entities within one single hyperedge, allowing one hyperedge to connect to many hyperedges and thereby rapidly increasing the branching width of the search.
Therefore, to control the knowledge retrieval process and selectively retrieve only the most relevant and necessary knowledge from the KH, we introduce the concept of reasoning plans.

\begin{definition}[Reasoning Plan]
Given a question $q$, a reasoning plan is a structured decomposition of $q$ represented as a pair $(\mathcal{Q}, \mathcal{L})$. 
Here, $\mathcal{Q} = \{q_1, q_2, \dots, q_m\}$ denotes the set of subquestions, where each $q_i$ addresses a partial aspect of $q$, and $\mathcal{L} \subseteq \mathcal{Q} \times \mathcal{Q}$ encodes dependency constraints among them. 
The relations $\mathcal{L}$ defines a partial order, that is, if $(q_i, q_j) \in \mathcal{L}$, then $q_i$ must be answered before $q_j$.
\end{definition}

\myparagraph{Question Subgraph Sketching}
While it is possible to generate reasoning plans directly from the internal knowledge of an LLM, such plans are often misaligned with the underlying KH. 
In particular, due to limited awareness of domain-specific relations, the LLM may introduce subquestions that cannot be resolved.
Nor is it sufficient to rely solely on the anchored graph elements, namely, topic entities $\mathcal{T}$ and target hyperedges $\mathcal{R}$, to form the plan context, as these elements do not reflect the broader relational structure required for multi-hop reasoning. 
To mitigate this issue, we construct a plan context graph that efficiently sketches the structure of the question subgraph. 
This is achieved by treating $\mathcal{T}$ and $\mathcal{R}$ as seeds for controlled exploration.
The resulting subgraph provides explicit grounding for plan generation and improves the alignment between the reasoning plan and the available knowledge.

\begin{definition}[Plan Context Graph]
Given a question $q$, 
the question subgraph $\mathcal{H}_q=(\mathcal{V}_q,\mathcal{E}_q)$, 
topic entity set $\mathcal{T}\subseteq\mathcal{V}_q$, 
target hyperedge set $\mathcal{R}\subseteq\mathcal{E}_q$, 
and a plan depth $d_p$, the plan context graph $\mathcal{H}_{p}=(\mathcal{V}_{p},\mathcal{E}_{p})$ is defined as a subgraph of $\mathcal{H}_q$, where $\mathcal{V}_{p}$ and $\mathcal{E}_{p}$ include entities and hyperedges that are highly relevant to $q$ and are within the $d_p$-hop neighborhood of either $\mathcal{T}$ or $\mathcal{R}$. 
\end{definition}

\noindent
We initialize the plan graph $\mathcal{H}_{p}=(\mathcal{V}_{p},\mathcal{E}_{p})$ using target hyperedges $\mathcal{R}$ and the hyperedges incident to the topic entities $\mathcal{T}$. 
These hyperedges also serve as the initial frontier for exploration. 
To guide the search direction, we employ a hyperedge scoring strategy. 
For each frontier hyperedge $e$, we first compute entity-level relevance scores for all entities $v \in e$ with respect to the question $q$:

\begin{equation}
    \mathrm{SE}(v \mid q) = \cos \left( \mathbf{z}(v^{\text{desc}}), \mathbf{z}(q) \right),
\label{eq:entityscore_emb}
\end{equation}
where $\mathbf{z}(\cdot)$ denotes the embedding function. 
For each neighboring hyperedge $e' \in \mathrm{Nbr}(e)$, we apply an aggregation function $\mathrm{AGG}(\cdot)$ to combine the scores of the overlapping entities $V(e) \cap V(e')$, producing a hyperedge-level relevance score with respect to $q$ and $e$. 
Details are provided in Appendix~\ref{appendix:alg:plan_init}.
\begin{equation}
    \mathrm{SH}(e' \mid q,~e) = \mathrm{AGG}\Bigl(\{\, \mathrm{SE}(v\mid q) \mid v \in V(e) \cap V(e') \,\}\Bigr).
\label{eq:hyperedgescore}
\end{equation}

\noindent
Based on the hyperedge-level relevance scores, low-scoring directions (hyperedges) are pruned, and the exploration will focus on directions that are supported by highly relevant entities.

\myparagraph{Initial Reasoning Plan Generation}
After constructing the plan context graph $\mathcal{H}_{p}$, we leverage it as a structured input for the LLM to propose reasoning plans. 
We transfer the graph structure into a natural language plan context by augmenting the hyperedges layer by layer, from the nearest neighborhood of the anchored graph elements to the more distant neighborhoods. 
This ensures that the context reflects local relevance and progressively broader structural information.
Formally, given the plan context graph $\mathcal{H}_{p}$, the plan context $c_p$ is defined as $c_p = \mathrm{FormPlanContext}(\mathcal{H}_{p}, \mathcal{T}, \mathcal{R})$,
where $\mathrm{FormPlanContext}$ denotes the procedure that extracts and formats the relevant subgraph structure into plan context.
The LLM is then prompted with the question $q$, the topic entities $\mathcal{T}$, and the plan context $c_p$ to generate initial reasoning plans.

\myparagraph{Initial Reasoning DAG construction}
Given a reasoning plan $(\mathcal{Q},\mathcal{L})$, a dependency $(q_i,q_j)\in\mathcal{L}$ is transitively covered if there exists $q_k\in\mathcal{Q}$ such that $(q_i,q_k)\in\mathcal{L}$ and $(q_k,q_j)\in\mathcal{L}$.
To obtain a minimal representation of $\mathcal{L}$, we apply a Hasse Reduction that removes all dependency relations that can be transitively covered. 
The resulting reduced relations $\mathcal{L}_H$ retains only immediate (non-transitive) dependencies between subquestions.
With $\mathcal{L}_H$, we formally define the reasoning DAG as follows.

\begin{definition}[Reasoning DAG]
A reasoning DAG is the graph abstraction of a reasoning plan $(\mathcal{Q}, \mathcal{L})$. 
It is defined as a directed acyclic graph (DAG) $D = (\mathcal{Q}, \mathcal{L}_H)$, where each node represents a subquestion $q_i \in \mathcal{Q}$ and each directed edge $(q_i, q_j) \in \mathcal{L}_H$ encodes the immediate dependency between $q_i$ and $q_j$. 
\end{definition}

\noindent
An example reasoning DAG is shown in Figure~\ref{fig:workflow}. For each initial reasoning plan, we construct a corresponding reasoning DAG and apply a topological sort on the reduced dependency relations $\mathcal{L}_H$ to obtain an execution order over the subquestions. This order will guide the level-by-level completion of the reasoning DAG. 
The processed reasoning DAGs form the initial reasoning DAG set $\mathcal{D}_0$. 

\subsection{Reasoning}\label{subsec:reasoning}

Once the initial reasoning DAGs are constructed, the next step is to query the KH under their guidance. 
More specifically, for a reasoning DAG $D$, questions at the first level without dependencies are resolved first.
The retrieved answers are used to refine the current reasoning DAG $D$, and the questions of the next level will be unlocked for reasoning.
This iterative process repeats until all subquestions are answered. The step answers to subquestions and supporting knowledge are stored in the completed reasoning DAG. We refer to this process of progressively resolving and refining DAGs as reasoning. 
Due to space limitations, the pseudo-code of the reasoning algorithm is provided in Appendix~\ref{appendix:alg:reasoning}.

\begin{definition}[Completed Reasoning DAG]
A completed reasoning DAG is defined as a DAG $D_{\mathrm{comp}}=(\mathcal{Q},\mathcal{L}_H,\mathcal{AP})$, where each node $q_j \in \mathcal{Q}$ is associated with a non-empty set of answer-path pairs $\mathcal{AP}[q_j]=\{(a_j,p_j)\}$. 
Here, $a_j$ is a candidate answer to $q_j$ and $p_j$ is a supporting reasoning path in $\mathcal{H}_q$. 
\end{definition}

\myparagraph{Reasoning as State-Space Search}
To explicitly model the reasoning process, we introduce the concept of a reasoning state.
A reasoning state records a reasoning DAG and the current progress in resolving its subquestions.
An initial state corresponds to an initial reasoning DAG $D \in \mathcal{D}_0$ with no subquestions resolved. A goal state corresponds to a completed reasoning DAG.

\begin{definition}[Reasoning State]
A reasoning state is a pair $(D,i)$, where $D=(\mathcal{Q},\mathcal{L}_H,\mathcal{AP})$ is a reasoning DAG and $i$ is the index of the current reasoning level. A transition from $(D,i)$ to $(D',i+1)$ occurs once all subquestions in the $i$-th level of $D$, denoted as $\mathcal{Q}_i$, are resolved with non-empty sets of answer-path pairs $\mathcal{AP}_i$.
\end{definition}

\noindent
Given the reasoning state defined above, we formulate the reasoning process as a structured search over reasoning states and solve it using Depth-First Search (DFS).
The frontier $\mathcal{F}$ is initialized with the set of initial reasoning DAGs $\mathcal{D}_0$. At each iteration, a state $(D,i)$ is popped from $\mathcal{F}$. 
If $D$ is incomplete, the subquestions $\mathcal{Q}_i$ at the current level $i$ are attempted. 
Each subquestion is resolved by retrieving reasoning paths from the anchored graph elements in the query subgraph $\mathcal{H}_q$, within a KH exploration depth limit $d_{max}$. Details of the retrieval procedure are provided in Section~\ref{subsec:stepanswer}. The retrieved answer-path pairs are then used to generate successor states. These successor states are pushed back into the frontier for later iterations. If $D$ is completed, it is added to the solution set $\mathcal{D}_{\mathrm{comp}}$. The search terminates once $|\mathcal{D}_{\mathrm{comp}}|\ge K$ or the frontier is empty, where $K$ is a configurable parameter denoting the maximum number of solutions.
Formally, the search procedure is summarized as $\mathcal{D}_{\mathrm{comp}} = \mathrm{Reasoning}(\mathcal{D}_0, \mathcal{H}_q, d_{\max}, b, K)$.

\myparagraph{State Transitions}
Given a state $(D,i)$, the sets of answer-path pairs $\mathcal{AP}_i$  at level $i$, we illustrate the state transition process as follows. Since each subquestion $q_j \in \mathcal{Q}_i$ is associated with a set of candidate assignments in the form of answer-path pairs $AP_j$, a valid joint assignment for level $i$ can be obtained by selecting one answer-path pair $(a_j, p_j) \in AP_j$ for every $q_j \in \mathcal{Q}_i$. 
If multiple joint assignments exist, the current state branches accordingly. 
For each joint assignment $\mathbb{AP}$, a successor reasoning DAG is constructed as $D_{\text{new}}=\mathrm{LLMGenerateNewDAG}\!\left(D,\,\mathbb{AP}\right)$.
Here, the LLM is prompted with the current set of subquestions $\mathcal{Q}$ together with one candidate answer for each completed subquestion. 
A refined reasoning plan is proposed by the LLM, which is then validated against the existing reasoning DAG $D$, and then merged with $D$ to produce the successor reasoning DAG $D_{\text{new}}$.

\subsection{Answer and Path Retrieval}\label{subsec:stepanswer}

The core subroutine for resolving subquestions is the retrieval of answer--path pairs from the question subgraph $\mathcal{H}_{q} = (\mathcal{V}_{q}, \mathcal{E}_{q})$. 
We employ beam search to progressively explore $\mathcal{H}_{q}$ with increasing depth.
At each depth level, the subquestion is attempted using the knowledge retrieved based on a reasoning path discovered so far.
The process terminates once a reasoning path provides sufficient knowledge to answer the subquestion. Due to space limitations, we detail the algorithm of answer and path retrieval in Appendix~\ref{appendix:alg:path}.

\myparagraph{Graph Re-anchoring}
For a subquestion $q_j$, the relevant knowledge should be within a more concise subregion of the question subgraph $\mathcal{H}_q$.
Accordingly, we re-anchor the graph, identify the topic entities $\mathcal{T}_j$ and target hyperedges $\mathcal{R}_j$ specific to $q_j$, following the same procedure described in Section~\ref{subsec:graph_achor}. 
$\mathcal{T}_j$ and $\mathcal{R}_j$ serve as seeds for the subsequent beam search, \ie we initialize the search frontier $\mathcal{F}_e$ with $\mathcal{R}_j$ and the hyperedges incident to $\mathcal{T}_j$.
The key challenge is then to guide the search toward discovering knowledge most relevant to the question $q_j$. 
To address this challenge, we design a strategy tailored to KHs.

\myparagraph{EWO-based Search Direction Selection}
In standard graphs, two adjacent edges can share at most a single node, in contrast, hyperedges in a hypergraph may overlap on multiple entities. 
Moreover, the contribution of these overlapping entities to answering a question is not uniform: some are irrelevant, while others provide critical evidence. 
Thus, neighboring hyperedges should neither be treated as equally relevant, nor should their relevance be determined solely by the number of shared entities. 
To address this, we propose the \textbf{Entity-Weighted Overlap (EWO)} score, a fine-grained strategy in which the relevance of a hyperedge is computed by aggregating the question-specific relevance of its overlapping entities $V(e) \cap V(e')$. 
We now describe the two-step procedure for computing EWO score.

\myparagraphunderline{Entity scoring}
Each overlapping entity $v$ is first assigned a provisional relevance score with respect to $q_j$ using the embedding-based similarity score defined in Eq.~\eqref{eq:entityscore_emb}. 
Entities with similarity scores above the threshold $\theta_{\text{emb}}$ are retained and further evaluated by the LLM to obtain a finer-grained relevance score. 
Entities with similarity scores below the threshold $\theta_{\text{emb}}$ are assigned a relevance score of $0$.
This score reflects how semantically relevant $v$ is for $q_j$.
\begin{equation}
\mathrm{EW}(v \mid q_j) \;=\;
\begin{cases}
  \mathrm{LLMScore}(v, q_j), & \text{if } \mathrm{SE}(v \mid q_j) \geq \theta_{\text{emb}}, \\[6pt]
  0, & \text{otherwise}.
\end{cases}
\end{equation}

\myparagraphunderline{Hyperedge scoring}
The hyperedge score integrates average and maximum scores over overlapping entities. The average score reflects overall semantic relevance, while the maximum score emphasizes salient entities providing strong evidence. This design captures both coarse-grained support and fine-grained discriminative cues.
\begin{equation}
\mathrm{EWO}(e' \mid q_j,~e) = \mathrm{AGG}\Bigl(\{\, \mathrm{EW}(v\mid q_j) \mid v \in V(e) \cap V(e') \,\}\Bigr).
\end{equation}

\noindent
With the EWO score, we now determine where to expand the search within the candidate search directions (partial reasoning paths) $F_{\text{cand}}$. 
In the first stage, $F_{\text{cand}}$ is ranked according to the hyperedge-level EWO scores of terminal hyperedges. 
From the resulting top-ranked directions, we prompt an LLM to select the top-$b$ directions $F_{\text{sel}} = \text{LLMSelectDirections}(F_{\text{cand}}, q_j, b)$. 
This evaluates the partial reasoning paths in context, rather than relying solely on the EWO score of the terminal hyperedge.

\myparagraph{Reasoning Path Selection}
At each depth, we construct a set of candidate reasoning paths $P_{\text{cand}}$ from the partial paths $\mathcal{P}$ explored so far. 
Each candidate reasoning path is then ranked using a path-level relevance score that aggregates the EW scores of entities along the path. 
Formally, the path-level score is defined as:

\begin{equation}
\mathrm{SP}(p) = \operatorname{sum}\{\mathrm{EW}(v \mid q_j) \;\mid\; v \in e \land e \in p \}
\end{equation}
\noindent
The top-ranked candidate reasoning paths are then passed to the LLM, to determine whether one or more of them provide sufficient knowledge to yield a step answer for $q_j$. The reasoning paths with sufficient knowledge are  $P_{\text{sel}} = \text{LLMSelectPaths}(P_{\text{cand}}, q_j)$.

\myparagraph{Step Answer for Subquestion}
Once $P_{\text{sel}}$ contains valid reasoning paths, we terminate the beam search and attempt to answer subquestion $q_j$, \ie the exploration beyond the current depth will not be performed.
For each selected path $p_j \in P_{\text{sel}}$, we construct the context for $q_j$ by combining three components: $i)$ the path itself, $ii)$ the descriptions of the entities covered by the path, and $iii)$ the originating text chunks of the hyperedges along the path.
Using the fused context $c_j$ as input, an LLM is invoked to produce the step answer $a_j = \text{LLMAnswerStep}(c_j, q_j)$.

\subsection{Final Answer Generation}\label{subsec:finalanswer}
As discussed in Section~\ref{subsec:reasoning}, the reasoning module produces a solution set $\mathcal{D}_{\text{comp}}$, where multiple reasoning plans (DAGs) are executed to completion with all sub-questions answered.
For each completed DAG $D_{\mathrm{comp}}=(\mathcal{Q},\mathcal{L}_H,\mathcal{AP})$, we aggregate the retrieved knowledge along its reasoning process following dependency relations $\mathcal{L}_H$ and uses this aggregated knowledge to generate a candidate answer $a(q)$ to the original question $q$.
An LLM-based evaluation agent is then introduced to assess these candidate answers $\mathcal{A}(q)$.
This judge evaluates each candidate answer $a(q)$ according to its consistency with the corresponding reasoning path, ultimately selecting the top-ranked answer as the final answer $a^*(q)$.

\begin{table*}[t]
\caption{\centering Results of \pokh across different domains, compared with the state-of-the-art (SOTA). The highest scores are highlighted in bold, while the second-best results are underlined for each dataset.}
\label{tab:mainresult}
\resizebox{\textwidth}{!}{
\begin{tabular}{@{}lccccccccccccccc@{}}
\toprule
\multirow{2}{*}{\textbf{Method}} & \multicolumn{3}{c}{\textbf{Medicine}} & \multicolumn{3}{c}{\textbf{Agriculture}} & \multicolumn{3}{c}{\textbf{CS}} & \multicolumn{3}{c}{\textbf{Legal}} & \multicolumn{3}{c}{\textbf{Mix}} \\ 
\cmidrule(lr){2-4} \cmidrule(lr){5-7} \cmidrule(lr){8-10} \cmidrule(lr){11-13} \cmidrule(lr){14-16}
 & F1 & R-S & G-E & F1 & R-S & G-E & F1 & R-S & G-E & F1 & R-S & G-E & F1 & R-S & G-E \\ \midrule
LLM-only & 12.89 & 0 & 43.27 & 12.74 & 0 & 46.85 & 18.65 & 0 & 48.87 & 21.64 & 0 & 49.05 & 16.93 & 0 & 45.65 \\
StandardRAG    & 27.90 & 62.57 & 55.66 & 27.43 & 45.81 & 57.10 & 28.93 & 48.40 & 56.89 & 37.34 & 51.68 & 59.97 & 43.20 & 47.26 & 64.62 \\
PathRAG        & 14.94 & 53.19 & 44.06 & 21.30 & 42.37 & 52.48 & 26.73 & 41.89 & 54.13 & 31.29 & 44.03 & 55.36 & 37.07 & 33.73 & 59.11 \\
HippoRAG2      & 21.34 & 59.52 & 49.57 & 12.63 & 18.58 & 44.85 & 17.34 & 23.99 & 47.87 & 18.53 & 34.42 & 45.93 & 21.53 & 18.42 & 46.35 \\
\hgr & 35.35 & 70.19 & 59.35 & 33.89 & \underline{62.27} & 59.79 & 31.30 & \underline{60.09} & 57.94 & 43.81 & 60.47 & 63.61 & 48.71 & \textbf{68.21} & 66.90 \\
\pokhl & 
\underline{45.63} & \underline{70.84} & \underline{59.90} & \underline{50.47} & \textbf{64.28} & \underline{63.13} & \underline{46.61} & \textbf{60.79} & \underline{60.17} & \underline{51.40} & \underline{64.58} & \underline{63.71} & \underline{53.81} & \underline{59.32} & \underline{61.32} \\
\rowcolor[HTML]{e7edf0}
\pokh & 
\textbf{52.94} & \textbf{74.02} & \textbf{67.35} & 
\textbf{56.67} & {58.88} & \textbf{69.46} & 
\textbf{54.15} & {57.72} & \textbf{66.79} & 
\textbf{58.81} & \textbf{65.22} & \textbf{69.88} & 
\textbf{69.16} & {59.86} & \textbf{76.17} \\

\bottomrule
\end{tabular}
}
\end{table*}

\section{Experiments}
\label{sec:exp}

\myparagraph{Experimental Settings} 
We evaluate \pokh on the KHQA datasets introduced in \hgr \cite{luo2025hypergraphrag}, which span five knowledge domains: {Medicine}, {Agriculture}, {CS}, {Legal}, and Mix.
Since the questions in the KHQA datasets are constructed from sampled knowledge fragments located only 1-3 hops away. 
We further extend these datasets with long-range questions to better assess the multi-hop reasoning capability of \pokh.
Specifically, we generate 200 additional questions per domain using knowledge fragments 3-6 hops away.
We also develop and evaluate \pokhl, a variant of \pokh that employs a fully embedding-based EWO score and uses only the hyperedges along the reasoning paths as context for answer generation. 
Following \cite{luo2025hypergraphrag}, we adopt three evaluation metrics: F1, Retrieval Similarity (R-S), and Generation Evaluation (G-E).
Due to the space limitation, experiment details, including 
baselines and implementation details, are provided in Appendix~\ref{appendix:exp_set}.

\subsection{Main Results}
\myparagraphquestion{(RQ1) Does \pokh outperform other methods}
As shown in Table~\ref{tab:mainresult}, \pokh achieves state-of-the-art performance across all domains in terms of F1 and G-E scores, outperforming the previous SOTA baseline \hgr by an average of 19.73\% and up to 22.85\% in F1 on the CS domain, as well as by an average of 8.41\% and up to 9.67\% in G-E.
For the R-S score, \pokh generally achieves comparable results to \hgr, with up to a 4.75\% improvement in the Legal domain. The main weakness appears in the Mix domain, which is reasonable since it integrates knowledge from multiple domains. Unlike \hgr, which prioritizes retrieving text with high semantic similarity, \pokh retrieves knowledge that contributes directly to reasoning toward the answer, even when such knowledge is not semantically similar to the surface context. Notably, the Mix domain is also where PathRAG, another reasoning-path-based retrieval method, attains its lowest R-S score, indicating a similar behavior pattern.
For the variant \pokhl, it surpasses \hgr by an average of 10.97\% and up to 16.58\% in F1 on the Agriculture domain. It also remains competitive with \hgr in terms of both G-E and R-S scores, showing up to 3.34\% improvement in the Agriculture domain and 4.11\% improvement in the Legal domain, respectively.

\newcolumntype{Y}{>{\centering\arraybackslash}X}
\begin{table}[t]
\centering
\caption{Ablation study results (F1).}
\label{tab:ablation}
\renewcommand{\arraystretch}{1.0}
\begin{tabularx}{\columnwidth}{lYYY}
\toprule
\textbf{Method} & \textbf{Agriculture} & \textbf{CS} & \textbf{Mix} \\
\midrule
\rowcolor[HTML]{e7edf0}
\pokh                  & 58.49 & 59.47 & 69.39 \\
w/o EWO Guide         & 53.22 & 56.27 & 65.63 \\
w/o Synonym Merge     & 53.26 & 54.96 & 64.74 \\
w/o Plan Context      & 53.70 & 53.67 & 64.59 \\
w/o Src Chunks        & 53.55 & 54.75 & 63.76 \\
w/o Target Hyperedge & 53.27 & 55.60 & 60.81 \\
w/o ALL               & 51.93 & 51.96 & 55.84 \\
\bottomrule
\end{tabularx}
\end{table}

\begin{table*}[!t]
\caption{
\centering Performance of \pokh and baselines on long-range multi-hop QA tasks across domains. The highest scores are highlighted in bold, while the second-best results are underlined for each dataset. }\label{tab:long_range}
\resizebox{\textwidth}{!}{
\begin{tabular}{@{}lccccccccccccccc@{}}
\toprule
\multirow{2}{*}{\textbf{Method}} & \multicolumn{3}{c}{\textbf{Medicine}} & \multicolumn{3}{c}{\textbf{Agriculture}} & \multicolumn{3}{c}{\textbf{CS}} & \multicolumn{3}{c}{\textbf{Legal}} & \multicolumn{3}{c}{\textbf{Mix}} \\ 
\cmidrule(lr){2-4} \cmidrule(lr){5-7} \cmidrule(lr){8-10} \cmidrule(lr){11-13} \cmidrule(lr){14-16}
 & F1 & R-S & G-E & F1 & R-S & G-E & F1 & R-S & G-E & F1 & R-S & G-E & F1 & R-S & G-E \\ 
\midrule
IO (LLM-only) & 23.36 & 0 & 0 & 30.95 & 0 & 0 & 39.27 & 0 & 0 & 43.20 & 0 & 0 & 48.98 & 0 & 0 \\
COT (LLM-only) & 22.70 & 0 & 43.42 & 34.32 & 0 & 52.22 & 42.41 & 0 & 58.16 & {43.36} & 0 & 57.71 & 50.48 & 0 & 62.24 \\
SC (LLM-only)  & 26.06 & 0 & 44.80 & 37.49 & 0 & 53.84 & {46.42} & 0 & {60.34} & 40.59 & 0 & 55.36 & 54.04 & 0 & 63.16 \\
StandardRAG & 21.90 & 66.02 & 50.13 & 48.00 & 50.92 & 67.81 & 27.34 & 59.42 & 57.11 & 35.86 & 55.25 & 60.37 & 52.20 & 59.79 & 69.91 \\
\hgr & {39.94} & \underline{71.72} & {60.81} 
& {52.40} & \underline{64.36} & {70.76} 
& 29.95 & \textbf{68.78} & {58.72} 
& {38.45} & {58.30} & {61.87} 
& \underline{64.55} & \textbf{70.23} & \underline{76.63} \\
\pokhl & \underline{47.30} & {71.24} & \underline{60.98} & \underline{70.02} & \textbf{65.75} & \underline{74.44} & \underline{62.11} & {65.09} & \underline{69.08} & \underline{60.25} & \underline{59.21} & \underline{67.78} & {61.19} & {62.48} & {66.16} \\
\rowcolor[HTML]{e7edf0}
\pokh & \textbf{51.26} & \textbf{75.11} & \textbf{65.23} 
& \textbf{81.01} & {62.22} & \textbf{84.18} 
& \textbf{74.83} & \underline{68.44} & \textbf{80.00} 
& \textbf{71.93} & \textbf{64.89} & \textbf{77.68} 
& \textbf{79.69} & \underline{68.45} & \textbf{82.25} \\ 
\bottomrule
\end{tabular}
}
\end{table*}
\subsection{Ablation Study}

\myparagraphquestion{(RQ2) Does the main component of \pokh work effectively}
We conduct an ablation study across three domains to quantify the contribution of each component. From each domain, Agriculture, CS, and Mix, we randomly sample 200 questions and report the F1 score for comparison. 
As shown in Table~\ref{tab:ablation}, removing the EWO Guide Direction Selection decreases F1 by up to 5.3\%, demonstrating its effectiveness in guiding the exploration of reasoning paths based on semantic flow from one hyperedge to another. 
Removing Synonym Merge reduces F1 by up to 5.2\%, indicating that this graph-reduction technique benefits planning and reasoning. 
Removing Plan Context results in a performance drop of up to 5.8\%, highlighting the importance of planning context for the feasibility and alignment of the initial reasoning plan.
Without Source Chunks, F1 declines by up to 5.6\%, suggesting that the source text provides additional context that contributes to more accurate answers. 
Eliminating Target Hyperedge Matching in graph anchoring yields the largest drop, up to 8.6\%, demonstrating the importance of leveraging semantics encoded in hyperedges. 
When all modules mentioned above are removed (w/o ALL), performance drops sharply by up to 13.6\%, indicating their complementary contributions.

\begin{figure}[t]
    \centering
    \begin{subfigure}[t]{0.48\columnwidth}
        \centering
        \includegraphics[width=\linewidth]{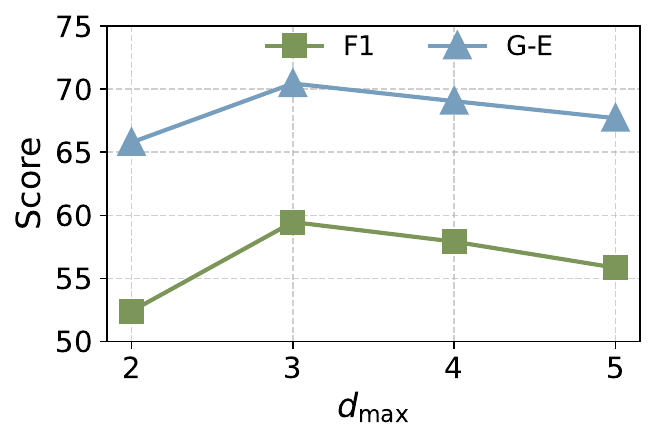}
        \caption{F1 \& G-E vs. $d_{\max}$}
        \label{fig:depth_a}
    \end{subfigure}
    \hfill
    \begin{subfigure}[t]{0.48\columnwidth}
        \centering
        \includegraphics[width=\linewidth]{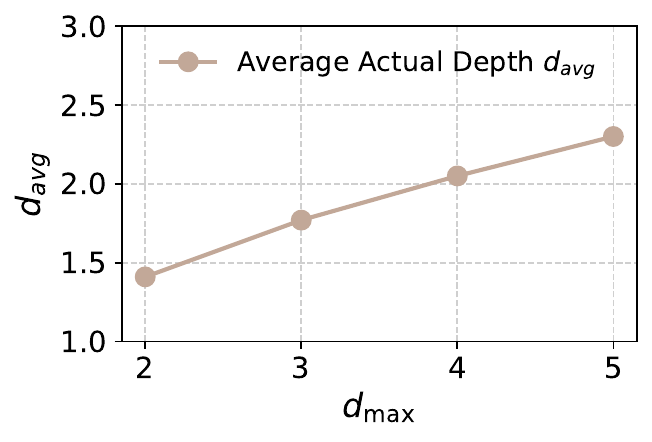}
        \caption{$d_{avg}$ vs. $d_{\max}$}
        \label{fig:depth_b}
    \end{subfigure}
    \caption{Impact of depth limit on performance. 
    }
    \label{fig:depth}
\end{figure}

\myparagraphquestion{(RQ3) How does the KH exploration depth affect the performance of \pokh} 
\pokh performs dynamic exploration on KH within the depth limit $d_{\max}$. To assess the impact of $d_{\max}$ on performance, we conduct experiments on 200 randomly sampled questions from the CS domain. 
We vary $d_{\max}$ from 2 to 5 and report the corresponding F1 and G-E scores in Figure~\ref{fig:depth}(a). We also report the actual exploration depth when the step answer for each subquestion is found and the exploration terminates in Figure~\ref{fig:depth}(b). 
As shown in Figure~\ref{fig:depth}(a), increasing $d_{\max}$ initially improves both F1 and G-E scores, with performance peaking at $d_{\max}$=3. 
Beyond this depth, a deeper search even introduces a slight degradation in both metrics.
This suggests that the additional search depth does not uncover additional useful information and instead introduces redundant context that fails to improve reasoning quality.
The trend in the actual exploration depth supports this interpretation. As $d_{\max}$ increases from 2 to 5, the average actual depth $d_{avg}$ grows modestly from 1.41 (at $d_{\max}$=2) to 2.30 (at $d_{\max}$=5). This indicates that \pokh rarely needs to utilize the full depth budget, as most subquestions are resolved within a relatively short reasoning path. 
This behavior can be attributed to two main reasons. 
$i)$ \pokh decomposes questions and dynamically refines its reasoning plan, thus simplifying each subquestion and shortening the reasoning path.
$ii)$ The EWO-guided exploration helps the system identify relevant paths early, minimizing unnecessary exploration.

\subsection{Effectiveness and Efficiency Evaluation}
\begin{figure}[t]
    \centering
    \includegraphics[width=1.0\linewidth]{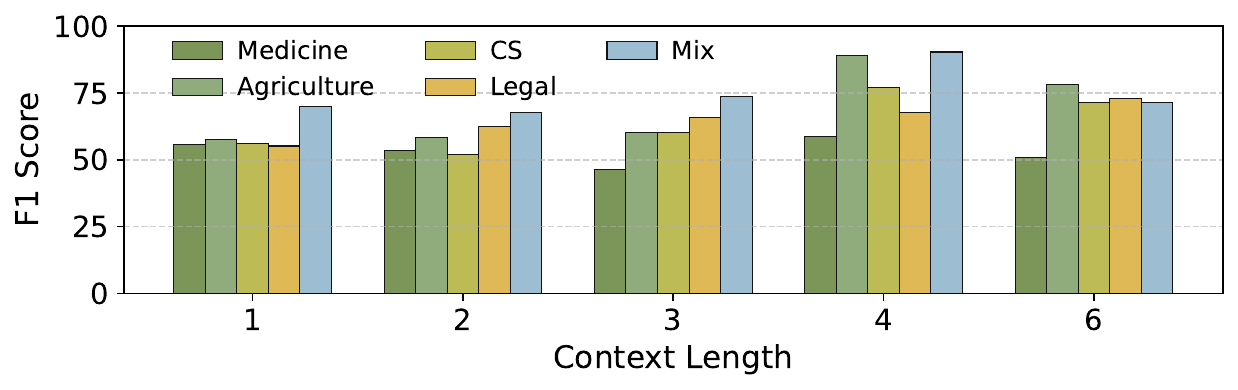}
    
    \caption{F1 score vs. length of ground-truth context. }\label{fig:context_length}

\end{figure}

\myparagraphquestion{(RQ4) Does \pokh stay effective on long-range multi-hop questions}
We further evaluate \pokh on 200 additional long-range questions per domain (3-6 hops). As shown in Table~\ref{tab:long_range}, \pokh sustains strong performance under these long-range settings, 
outperforming \hgr by an average of 26.68\% and up to 44.87\% in F1 in the CS domain, as well as by an average of 12.11\% and up to 21.28\% in G-E.
For the R-S score, \pokh achieves an average of 1.14\% and up to 6.59\% improvement in the Legal domain.
The variant \pokhl, also demonstrates strong performance under these long-range settings, outperforming \hgr by an average of 15.11\% and up to 32.15\% in F1.
The robustness of \pokh is also supported by the results in Figure~\ref{fig:context_length}, which analyzes the effect of ground truth context length (the number of knowledge fragments used when the question and golden answer are generated) on \pokh’s performance. The F1 scores remain stable as the context length grows, 
suggesting that \pokh can maintain reasoning coherence even when the relevant knowledge spans multiple hops, demonstrating its robustness in long-range, multi-hop reasoning.

\newcolumntype{Y}{>{\centering\arraybackslash}X}
\newcolumntype{M}[1]{>{\centering\arraybackslash}m{#1}} % centred + vertical middle
\begin{table}[t]
\centering
\caption{\#Token per question across domains.}
\label{tab:tokens_per_question}
\renewcommand{\arraystretch}{1.0}
\begin{tabularx}{\columnwidth}{
    l
    >{\hsize=1.7\hsize}Y
    >{\hsize=0.9\hsize}Y
    >{\hsize=0.9\hsize}Y
    >{\hsize=0.5\hsize}Y
}
\toprule
\textbf{Domain} & \textbf{\hgr} & \textbf{PRoH-L} & \textbf{\#Token\%$\downarrow$} & \textbf{F1$\uparrow$} \\
\midrule
Medicine    & 21,112 & 19,732 & $6.54\%$  & 10.28 \\
Agriculture & 17,914 & 12,528 & $30.07\%$ & \textbf{16.58} \\
CS          & 18,666 & 12,166 & $\mathbf{34.82\%}$ & 15.31 \\
Legal       & 22,086 & 28,831 & $-30.54\%$ & 7.59 \\
Mix         & 13,856 & 9,687  & $30.09\%$ & 5.10 \\
\bottomrule
\end{tabularx}
\end{table}

\myparagraphquestion{(RQ5) Is \pokhl cost efficient in token consumption}
As shown in Table~\ref{tab:tokens_per_question}, \pokhl demonstrates notable efficiency in token usage while maintaining competitive performance across all domains. Compared with \hgr, \pokhl significantly reduces the number of tokens used (input + output) per question, with the largest savings observed in the CS domain at 34.82\%. Despite the reduced token usage, \pokhl achieves consistent improvements in F1 score with up to 16.58\% in the Agriculture domain. The only exception appears in the Legal domain, where token consumption increases for \pokhl, however, this increase still yields a positive F1 gain of 7.59\%. Overall, these results confirm that \pokhl achieves a superior balance between efficiency and accuracy, offering a cost-effective alternative to full \pokh.

\myparagraphquestion{(RQ6) How does the performance of \pokh vary across different LLM backbones}
To study the impact of LLM backbones, we evaluate our method using LLMs of varying model sizes on the long-range questions from the Agriculture and CS domains.
As reported in Table~\ref{tab:llm_backbone}, stronger LLMs consistently improve F1 and G-E scores across both domains. As Qwen3 scales from 4B to 14B to 80B, the F1 score increases from 78.10\% to 79.43\% to 84.27\% in Agriculture and from 67.12\% to 73.48 to 75.97\% in CS. This trend indicates that larger models are more effective at planning and retrieving question-relevant knowledge from KHs, enabling more effective evidence integration for high-quality answer generation.

\vspace{5mm}

\textit{
To further evaluate \pokh, we perform additional ablation studies on long-range multi-hop question answering and examine the effects of the state search strategy and planning depth. We also analyze token usage across modules. 
Detailed results are provided in Appendix~\ref{appendix:exp}.
}

\newcolumntype{Y}{>{\centering\arraybackslash}X}

\begin{table}[t]
\centering
\caption{Performance across LLM backbones.}
\label{tab:llm_backbone}
\renewcommand{\arraystretch}{1.0}
\begin{tabularx}{\columnwidth}{lYYYYYY}
\toprule
\multirow{2}{*}{\textbf{LLM backbone}} 
& \multicolumn{3}{c}{\textbf{Agriculture}} & \multicolumn{3}{c}{\textbf{CS}} \\
\cmidrule(lr){2-4} \cmidrule(lr){5-7}
& F1 & R-S & G-E
& F1 & R-S & G-E \\
\midrule
Qwen3-4B  & 78.10 & 63.89 & 82.26 & 67.12 & 67.50 & 74.03 \\
Qwen3-14B & 79.43 & \textbf{64.49} & 83.41 & 73.48 & 67.95 & 79.16 \\
Qwen3-80B & \textbf{84.27} & 62.97 & \textbf{86.42} & \textbf{75.97} & \textbf{69.07} & \textbf{81.87} \\
\bottomrule
\end{tabularx}
\end{table}

\section{Conclusion}
\label{sec:conclusion}

This paper presents \pokh, a dynamic Knowledge Hypergraph-based RAG framework for multi-hop question answering. 
By introducing context-aware planning, structured iterative question decomposition and an EWO-guided reasoning path retrieval strategy, \pokh enables adaptive planning and reasoning on knowledge hypergraphs with beyond binary relational structures. Experimental results demonstrate that \pokh achieves SOTA performance across multiple knowledge domains, surpassing the prior SOTA by an average of 19.73\% in F1 and 8.41\% in G-E score, while maintaining high robustness in long-range multi-hop reasoning tasks.

\begin{acks}
Xiaoyang Wang is supported by the Australian Research Council DP230101445 and DP240101322. Wenjie Zhang is supported by the Australian Research Council DP230101445 and FT210100303.
\end{acks}

\bibliographystyle{ACM-Reference-Format}

\bibliography{acmart}

% \renewcommand{\theHsection}{A\arabic{section}}
% \appendixsiam
% \clearpage

\appendix
\section{Algorithm}\label{appendix:alg}

\subsection{Planning}\label{appendix:alg:plan_init}
We summarize the comprehensive algorithmic procedure of planning as shown in Algorithm \ref{alg:plan_init}.

\begin{algorithm}[h]
\small
\SetVline
\caption{PlanInit}\label{alg:plan_init}
\Input{Question Subgraph $\mathcal{H}_{q}=(\mathcal{V}_{q},\mathcal{E}_{q})$, Question $q$, Topic Entities $\mathcal{T}$, Target Hyperedges $\mathcal{R}$, Plan Depth $d_p$, \#Initial Plans $n_0$}
\Output{Set of initial DAGs $\mathcal{D}_{0}$}

\State{$\mathcal{E}_{p} \gets \{e \mid v \in e \land v \in \mathcal{T}\} \cup \mathcal{R}$}
\State{$\mathcal{V}_{p} \gets \{v \mid v \in e \land e \in \mathcal{E}_{p}\}$}

\State{$\mathcal{F}_e \gets \{e \mid v \in e \land v \in \mathcal{T}\} \cup \mathcal{R}$}

\For{$d \gets 1$ \KwTo $d_{p}$}{
\State{$\mathcal{F}_e' \gets \emptyset$}

  \ForEach{$e \in \mathcal{F}_e$}{

    \State{$F_{\text{cand}} \gets \emptyset$}

    \ForEach{$e'  \in \mathrm{Nbr}(e)$}{
       
        \State{$S_{\max} \gets \max \{ \mathrm{SE}(v \mid q) \mid v \in V(e) \cap V(e') \}$}
        \State{$S_{\mathrm{avg}} \gets \operatorname{mean}\{\mathrm{SE}(v \mid q) \mid v \in V(e) \cap V(e')\}$}
        \State{$\mathrm{SH}(e' \mid q, e) \gets \mathrm{Combine}(S_{\max}, S_{\mathrm{avg}})$}       
        \State{$F_{\text{cand}} \gets F_{\text{cand}} \cup \{(e', \mathrm{SH}(e' \mid q, e))\}$}
        
    }
    \State{$F_{\text{sel}} \gets \text{RankSelectDirections}(F_{\text{cand}})$}

    \State{$\mathcal{F}_e' \gets \mathcal{F}_e' \cup F_{\text{sel}}$}
    \State{$\mathcal{E}_{p} \gets \mathcal{E}_{p} \cup F_{\text{sel}}$}
    \State{$\mathcal{V}_{p} \gets \mathcal{V}_{p} \cup \{v \mid v \in e \land e \in F_{\text{sel}}\}$}

  }
  \State{$\mathcal{F}_e \gets \mathcal{F}_e'$}
}

\State{$c_p \gets \text{FormPlanContext}(\mathcal{H}_{p})$}
\State{$\mathcal{D}_{0} \gets \emptyset$}

\For{$i \gets 1$ \KwTo $n_{0}$}{
\State{$\mathcal{Q},~\mathcal{L} \gets $LLMPlan$(q, \mathcal{T}, c_p)$ }
\State{$D \gets $HasseTopSort$(\mathcal{Q},~\mathcal{L})$}
\State{$\mathcal{D}_{0} \gets \mathcal{D}_{0} \cup \{D\}$}

}
\Return $\mathcal{D}_{0}$\;

\end{algorithm}
\newpage
\subsection{Reasoning}\label{appendix:alg:reasoning}
We summarize the comprehensive algorithmic procedure of reasoning as shown in Algorithm \ref{alg:reasoning}.
\begin{algorithm}[h]
\small
\SetVline
\caption{Reasoning}\label{alg:reasoning}
\Input{Initial DAGs $\mathcal{D}_{0}$, Question Subgraph $\mathcal{H}_{q}$, KH Exploration Depth Limit $d_{\max}$, Beam Width $b$, Max \#Solutions $K$}
\Output{Completed DAGs $\mathcal{D}_{\text{comp}}$}

\State{$\mathcal{D}_{\text{comp}} \gets \emptyset$}
\State{$\mathcal{F} \gets \emptyset$}

\ForEach{$D\in \mathcal{D}_{0}$}{
\State{$\mathcal{F} \gets \mathcal{F} \cup \{(D, 0)\}$}
}

\While{$\mathcal{F} \neq \emptyset$ \textbf{and} $|\mathcal{D}_{\text{comp}}| < K$}{
  \State{$(D,i) \gets \mathcal{F}.pop()$}

  \If{$i \ge |D.\text{levels}|$}{
    \State{$\mathcal{D}_{\text{comp}} \gets \mathcal{D}_{\text{comp}} \cup \{D\}$}
    \State{\textbf{continue}}
  }

  \State{$\mathcal{Q}_{i} \gets$ subquestions at level $i$ of $D$}

  \State{$\mathcal{AP}_{i} \gets \emptyset$}

  \ForEach{$q_j \in \mathcal{Q}_{i}$}{
    $AP_j \gets $RetrieveAnswersWithPaths$(\mathcal{H}_{q}, q_j, d_{\max}, b)$\;

    $\mathcal{AP}_{i}[j] \gets AP_j$\;
  }

  \ForEach{combination of answers $\mathbb{AP}$ in $\mathcal{AP}_{i}$}{
  
    \State{$D_{\text{new}} \gets $LLMGenerateNewDAG$(D, \mathbb{AP})$}

    \State{$\mathcal{F}.push((D_{\text{new}}, i+1))$}
  
  }
}

\Return $\mathcal{D}_{\text{comp}}$\;
\end{algorithm}

\newpage
\subsection{Answer and Path Retrieval}\label{appendix:alg:path}
We summarize the comprehensive algorithmic procedure of answer and path retrieval as shown in Algorithm \ref{alg:retrieve_answers_with_paths_hl}.

\begin{algorithm}[h]
\small
\SetVline
\caption{RetrieveAnswersWithPaths}\label{alg:retrieve_answers_with_paths_hl}
\Input{Question Subgraph $\mathcal{H}_{q}=(\mathcal{V}_{q},\mathcal{E}_{q})$, Subquestion $q_j$, KH Exploration Depth Limit $d_{\max}$, Beam Width $b$}
\Output{Set of Step Answer - Reasoning Path Pair $AP_j$}

\tcp{Graph re-anchoring}
\State{$\mathcal{T}_j \leftarrow $TopicEntityInit$(q_j, \mathcal{H}_q)$}
\State{$\mathcal{R}_j \leftarrow $TargetHyperedgeMatch$(q_j, \mathcal{H}_q)$}

\tcp{Initialize frontier}
\State{$\mathcal{F}_e \gets \emptyset$}
\State{$\mathcal{P} \gets \emptyset$}
\ForEach{$e \in \{e \mid v \in e \land v \in \mathcal{T}_j\} \cup \mathcal{R}_j$} {
\State{$\mathcal{F}_e \gets \mathcal{F}_e \cup \{(e,[e])\}$}

\State{$\mathcal{P} \gets \mathcal{P}\cup \{[e]\}$}

}

\For{$d \gets 1$ \KwTo $d_{\max}$}{

  \tcp{Beam search from current frontier}

  \State{$F_{\text{cand}} \gets \emptyset$}
  
  \ForEach{$(e, p_e)  \in \mathcal{F}_e$}{

    \ForEach{$e'  \in \mathrm{Nbr}(e)$}{
        
        \State{$S_{\max} \gets \max \{ \mathrm{EW}(v \mid q_j) \mid v \in V(e) \cap V(e') \}$}
        \State{$S_{\mathrm{avg}} \gets \operatorname{mean}\{\mathrm{EW}(v \mid q_j) \mid v \in V(e) \cap V(e')\}$}
        \State{$\mathrm{EWO}(e' \mid q_j, e) \gets \mathrm{Combine}(S_{\max}, S_{\mathrm{avg}})$}    
        
        \State{$p_{e'} \gets p_e \oplus [e']$}

        \State{$F_{\text{cand}} \gets F_{\text{cand}} \cup \{(p_{e'}, \mathrm{EWO}(e' \mid q_j, e))\}$}
        
    }
  }
  \State{$F_{\text{cand}} \gets \text{RankSelectDirections}(F_{\text{cand}})$}
  \State{$F_{\text{sel}} \gets \text{LLMSelectDirections}(F_{\text{cand}}, q_j, b)$}
  \State{update $\mathcal{F}_e$ and $\mathcal{P}$ with $F_{\text{sel}}$}

  \tcp{Form and select reasoning paths}

  \State{$P_{\text{cand}} \gets \text{FormPaths}(\mathcal{P})$}
  \State{$P_{\text{cand}} \gets \text{RankSelectPaths}(P_{\text{cand}})$}
  \State{$P_{\text{sel}} \gets \text{LLMSelectPaths}(P_{\text{cand}}, q_j)$}

  \If{$P_{\text{sel}} \neq \emptyset$}{
    \tcp{Attempt subquestion}

    \State{$AP_j \gets \emptyset$}
    \ForEach{$p_j \in P_{\text{sel}}$}{
    
    \State{$c_j \gets \text{KnowledgeFusion}(p_j)$}
    \State{$a_j \gets \text{LLMAnswerStep}(c_j, q_j)$}

    \State{$AP_j \gets AP_j \cup \{(a_j, p_j)\}$}
    }

    \Return $AP_j$;
    
  }

}
\Return $\emptyset$\;
\end{algorithm}

\newcolumntype{Y}{>{\centering\arraybackslash}X}

\begin{table}[h]
\centering
\caption{Ablation study results on long-range multi-hop QA.}
\label{tab:ablation_module}
\renewcommand{\arraystretch}{1}
\begin{tabularx}{\columnwidth}{lYYYY}
\toprule
\multirow{2}{*}{\textbf{Method}} 
& \multicolumn{2}{c}{\textbf{Agriculture}} 
& \multicolumn{2}{c}{\textbf{Legal}} \\
\cmidrule(lr){2-3} \cmidrule(lr){4-5}
& F1 & G-E 
& F1 & G-E  \\
\midrule
\rowcolor[HTML]{e7edf0}
PRoH              & 81.01 & 84.18 & 71.93 & 77.68 \\
w/o Planning       & 78.28 & 82.36 & 68.49 & 75.36 \\
w/o Reasoning      & 78.12 & 82.39 & 71.13 & 76.48 \\
w/o Path retrieval & 69.55 & 75.95 & 65.06 & 71.74 \\
w/o Retrieval      & 30.61 & 49.12 & 33.24 & 50.77 \\
\bottomrule
\end{tabularx}
\end{table}

\newpage
\section{Additional Experiment}\label{appendix:exp}

\subsection{Ablation Study}

\myparagraphquestion{(RQ7) How do individual PRoH modules affect long-range multi-hop QA performance}
We conduct an ablation study on additional long-range questions from the Agriculture and Legal domains to quantify the contribution of individual modules.
As shown in Table~\ref{tab:ablation_module}, 
removing the planning module leads to an F1 drop of up to 3.44\%, indicating that context-aware structured question decomposition is necessary to divide and conquer long-range questions. Excluding the reasoning module results in an F1 decrease of up to 2.89\%, suggesting that iterative refinement over reasoning plans helps correct early-stage errors. Disabling path retrieval and only retrieving anchored graph elements (shallow retrieval) causes a noticeable performance drop of up to 11.46\% in F1, highlighting the importance of deep exploration over KHs to capture multi-hop relational chains for long-range QA. When the retrieval module is removed entirely, F1 drops substantially by up to 50.4\%, indicating a performance collapse. Overall, these results demonstrate that each module contributes effectively to PRoH’s multi-hop QA capacity.

\newcolumntype{Y}{>{\centering\arraybackslash}X}

\begin{table}[t]
\centering
\caption{Performance vs. state search strategy.}
\label{tab:tree_stat}
\renewcommand{\arraystretch}{1.0}

\begin{tabularx}{\columnwidth}{ccYYY YYY}
\toprule 
\multirow{2}{*}{\makecell{\textbf{\# Init}\\ \boldmath$n_0$}} &
\multirow{2}{*}{\makecell{\textbf{\# Soln}\\ \boldmath$K$}} &
\multicolumn{3}{c}{\textbf{BFS}} & \multicolumn{3}{c}{\textbf{DFS}} \\
\cmidrule(lr){3-5} \cmidrule(lr){6-8}
& & F1 & $\hat{w}_T$ & $V_T$ & F1 & $\hat{w}_T$ & $V_T$ \\
\midrule
1 & 1 & 57.86 & 28.00  & 8.31  & 55.52 & 11.83 & 3.29 \\
1 & 2 & 56.93 & 41.37  & 9.70  & 53.22 & 12.88 & 4.50 \\
1 & 3 & 57.70 & 18.58  & 9.24  & 57.16 & 12.65 & 5.67 \\
2 & 2 & 54.80 & 44.62  & 14.02 & 54.98 & 10.21 & 4.77 \\
2 & 3 & 58.58 & 39.13  & 15.15 & 57.15 & 10.04 & 6.24 \\
3 & 3 & 61.94 & 126.85 & 20.56 & 58.48 & 12.25 & 6.58 \\
\bottomrule
\end{tabularx}
\end{table}

\myparagraphquestion{(RQ8) How does the state search strategy affect the performance of \pokh}
For this experiment, we randomly sample 200 questions from the Medicine domain, and compare breadth-first (BFS) and depth-first (DFS) state search strategies under different settings of the number of initial plans $n_0$ and the maximum number of solutions $K$.
As reported in Table~\ref{tab:tree_stat}, the BFS strategy consistently achieves higher F1 than the DFS strategy, however, this performance advantage comes with a significant extra computational cost. When $n_0$=3 and $K$=3, BFS exhibits explosive growth in search width $\hat{w}_T$, it also visits 20.56 states on average, which is more than 3x of DFS.
DFS, though, as expected, has a much more stable width.
When we fix one of $n_0$ or $K$, increasing the other will always improve the F1 score for both strategies.
Overall, DFS offers a better performance-to-cost ratio and shows a more stable scaling behavior.

\myparagraphquestion{(RQ9) How does the plan depth affect the state search tree} As shown in Table~\ref{tab:plandepth}, increasing the plan depth consistently improves both F1 and G-E scores. When $d_p$ increases from $1$ to $3$, F1 and G-E rise from 55.65\% to 59.47\% and from 67.71\% to 70.45\%, respectively. This indicates that deeper plan depth provides more comprehensive context for planning.
The average peak search tree depth $\hat{d}_T$ increases when $d_p$ increases from $1$ to $2$ and then decreases to 1.36 when $d_p$=3, suggesting that, with a richer planning context, a more efficient plan can be generated.
Overall, deeper plan depth $d_p$ enhances performance without introducing excessive reasoning complexity through question decomposition.

\newcolumntype{Y}{>{\centering\arraybackslash}X}

\begin{table}[t]
\centering
\caption{Performance vs. plan depth.}
\label{tab:plandepth}
\renewcommand{\arraystretch}{1.0}
\begin{tabularx}{\columnwidth}{l c YYY}
\toprule
\multicolumn{2}{c}{\textbf{Plan depth \boldmath$d_p$}} & \textbf{F1} & \textbf{G-E} & \boldmath$\hat{d}_T$ \\
\midrule
& 1 & 55.65 & 67.71 & 1.38 \\
& 2 & 57.13 & 68.57 & 1.42 \\
& 3 & 59.47 & 70.45 & 1.36 \\
\bottomrule
\end{tabularx}
\end{table}

\subsection{Effectiveness and Efficiency Evaluation}

\begin{figure}[h]
    \centering
    \includegraphics[width=1.0\linewidth]{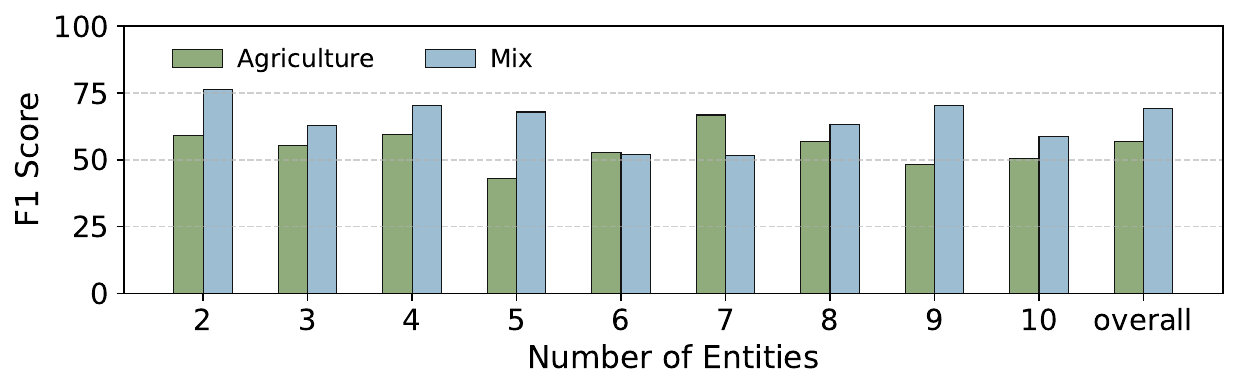}
    
    \caption{F1 score vs. \#entities in ground-truth context.}\label{fig:nary}
    % \vspace{5mm}
    
\end{figure}

\myparagraphquestion{(RQ10) How does \pokh perform with multiple entities in the ground-truth context} 
To evaluate the robustness \pokh to question complexity with respect to entity cardinality, we categorize questions from the Agriculture and Mix domains by the number of entities appearing in the ground-truth context and report the F1 score across varying entity cardinality.
Figure~\ref{fig:nary} reports the average F1 scores of \pokh across different levels of relational complexity in the questions. The complexity is measured by the number of entities participating in the ground-truth context for the question. 
Overall, \pokh exhibits stable performance as the number of entities increases, indicating robustness to growing relational complexity.

\begin{figure}[h]
    \centering
    \includegraphics[width=1.0\linewidth]{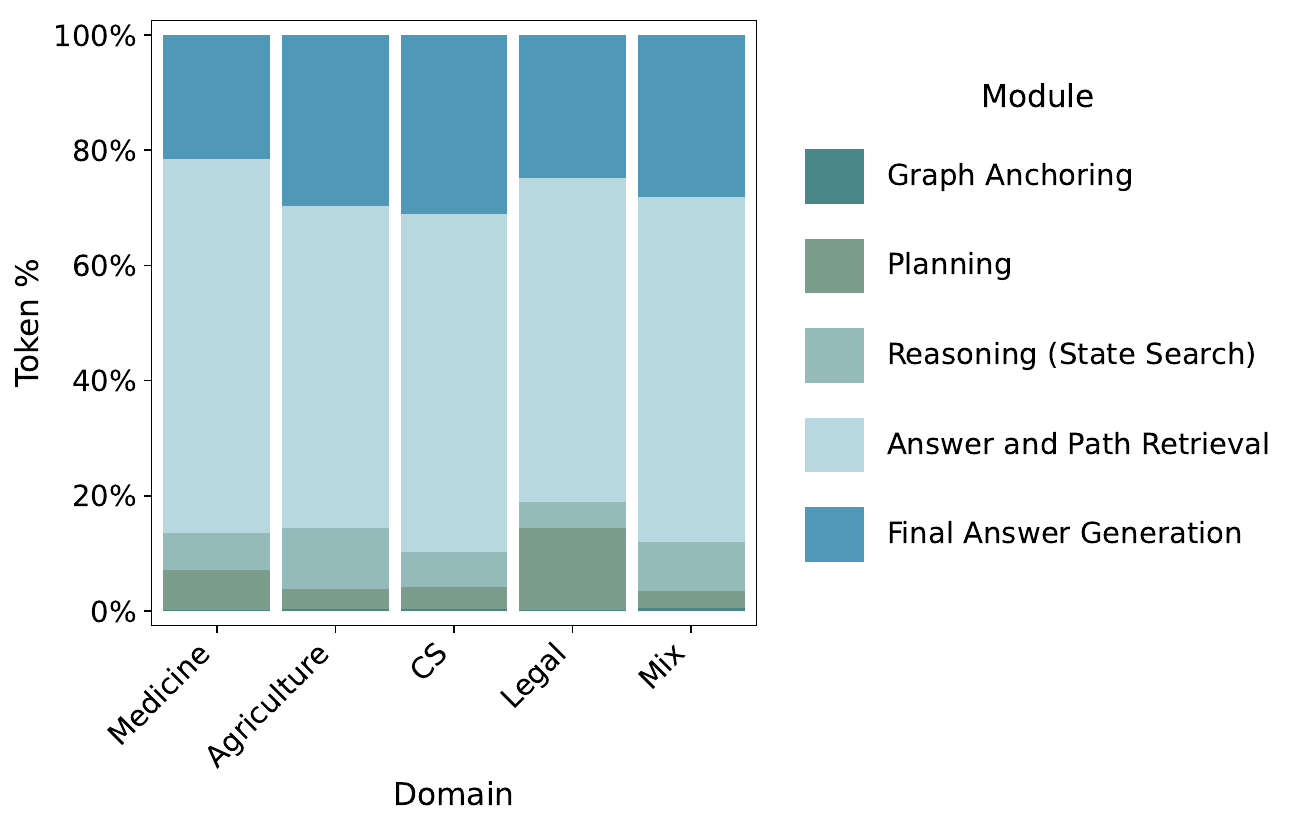}
    \caption{Token usage among modules.}\label{fig:token_dist}
    % \vspace{5mm}
\end{figure}

\myparagraphquestion{(RQ11) How does token usage distribute across modules}
Figure~\ref{fig:token_dist} summarizes token consumption across five domains, revealing a highly skewed distribution toward the Answer and Path Retrieval module. This is largely attributable to the high token overhead of LLM-based entity scoring and path selection. Final Answer Generation remains secondary across domains, where retrieved knowledge from multiple completed reasoning DAGs is leveraged for answer generation and selection. Planning consistently ranks third, reflecting the moderate overhead of conditioning on planning context. Reasoning (\ie DAG refinement) consumes relatively few tokens, accounting for only a small fraction of overall token usage.

% \newpage

\section{Experiment Details}\label{appendix:exp_set}

\myparagraph{Baselines}
We compare \pokh against baselines in four categories.
$i)$ \textbf{LLM-only}: Directly generates answers using the intrinsic knowledge of the LLM; 
$ii)$ \textbf{StandardRAG}: A conventional chunk-based RAG approach; 
$iii)$ \textbf{Graph-based RAG}: Graphs are a natural representation for modeling relational structure among entities~\cite{DBLP:conf/pakdd/HeYWZYYZ25, zhu2025unicom, zhai2025graph, zhai2025sgpt}, we include two graph-based RAG methods as baselines, namely {PathRAG}~\cite{DBLP:journals/corr/abs-2502-14902} and {HippoRAG2}~\cite{jimenez2025hipporag2}; 
$iv)$ \textbf{Knowledge Hypergraph-based RAG}: Hypergraphs are widely adopted to model higher-order relationships~\cite{pan2025hitec, li2024hypergraph, li2025dhg, xie2025hl}, we include the state-of-the-art KH-based RAG method {\hgr}~\cite{luo2025hypergraphrag} as baseline. 
As \hgr is the current SOTA, we directly refer to the results reported in their paper for \hgr and other baselines in our comparisons.

\myparagraph{Experimental Settings} 
Experiments are conducted using GPT-4o-mini as the primary LLM backbone, and text-embedding-3-small for vector embedding.
For \pokh, we set plan depth $d_p = 3$,  KH exploration depth limit $d_{max} = 3$, number of initial plans $n_0 = 2$, max number of solutions $K=2$.
For \pokhl, we set plan depth $d_p = 2$,  KH exploration depth limit $d_{max} = 3$, number of initial plans $n_0 = 1$, max number of solutions $K=1$. 

\textit{The complete implementation of \pokh, including prompt configurations, is publicly available.\footnote{\url{https://github.com/zaixjun/PRoH}.}}

\section{Case Studies and Error Analysis}\label{appendix:case_study}

% In this section, we present an analysis of error origins, followed by case studies illustrating both successful and representative error cases.
In this section, we analyze error origins and present case studies illustrating \pokh's structured question decomposition mechanism and representative failure cases.

\subsection{Error Analysis}
\begin{figure}[h]
    \centering
    \includegraphics[width=1.0\linewidth]{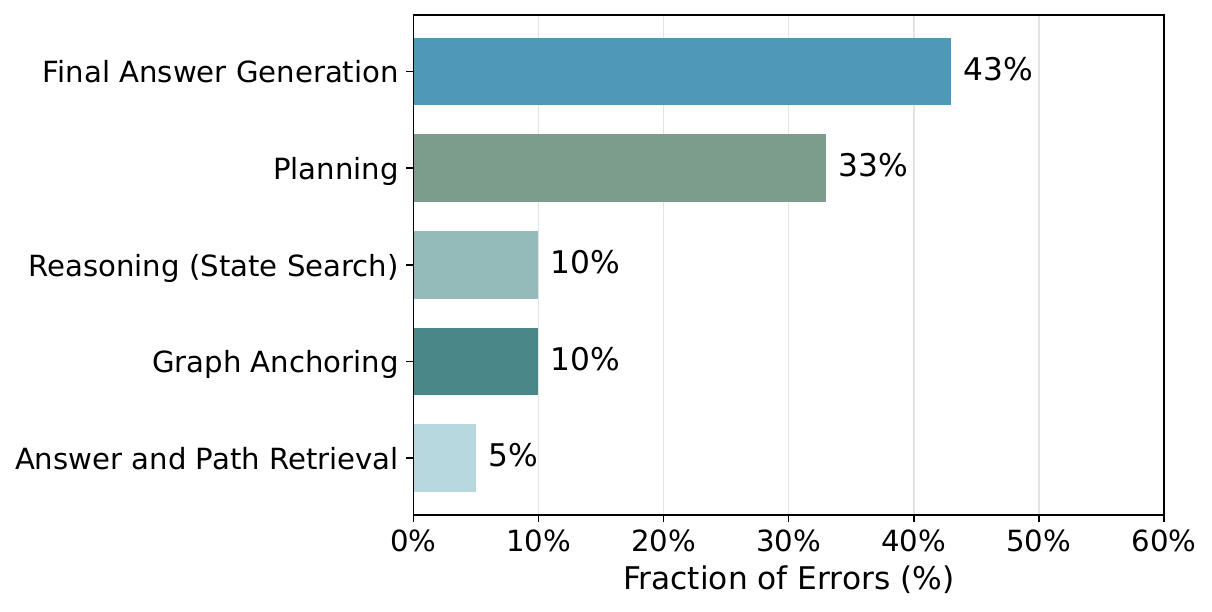}
    \caption{Error attribution by module.}\label{fig:error_origin}
    \vspace{5mm}
\end{figure}

% \myparagraphquestion{(RQ12) From which module do errors originate}
To determine where errors originate, we study the error distribution across modules and report their respective contributions in Figure~\ref{fig:error_origin}.
The results show a clear dominance of the Final Answer Generation module (43\%), where the model must integrate dispersed evidence. Planning emerges as the second-largest contributor (33\%), indicating that structured question decomposition and subgoal formulation remain challenging. Errors associated with Graph Anchoring and Reasoning (State Search) are comparatively infrequent. Answer and Path Retrieval exhibits the smallest error proportion, indicating that the retrieval for decomposed subquestions is relatively stable.

\clearpage
\onecolumn

\subsection{Example of Structured Question Decomposition}
Table~\ref{tab:case_gaap} illustrates \pokh's structured question decomposition mechanism. 
It also demonstrates \pokh's effectiveness in handling multi-entity and multi-hop question answering tasks.

\begin{table*}[h]
\centering
\caption{Example of Planning and Reasoning.}
\label{tab:case_gaap}
% \vspace{-3mm}
\begin{tabularx}{\textwidth}{@{}p{3cm}X@{}}
\toprule
\textbf{Field} & \textbf{Content} \\
\midrule
\textbf{Question} & What must be prepared in accordance with GAAP for financial and tax reporting purposes? \\

\textbf{Golden Answer} & {FINANCIAL STATEMENTS} \\
\textbf{Context} &
(1) The books of the Partnership shall be maintained, for financial and tax reporting purposes, on an accrual basis in accordance with GAAP. \par
(2) 1ACCOUNTING AND OTHER TERMS \par
(3) all accounting terms used herein shall be interpreted, all accounting determinations hereunder shall be made, and all {financial statements} required to be delivered hereunder shall be prepared in accordance with GAAP as in effect from time to time. \\
\textbf{nary} & 2 \\
\textbf{nhop} & 3 \\

\midrule
\textbf{DAG Edges} & $0 \rightarrow 1$, $0 \rightarrow 2$ \\
% \midrule

\textbf{Subquestion 0} &

\textbf{Subquestion:} What does GAAP stand for? \par
\textbf{Topics:} \verb|"GAAP"| \par
\textbf{Level:} 0 \par
\textbf{Answer:} Generally Accepted Accounting Principles \par

\textbf{Reasoning Path:} \texttt{\textless hyperedge\textgreater "“GAAP” means U.S. generally accepted accounting principles."} \\
% \midrule
\textbf{Subquestion 1} &
% \textbf{ID:} 1 \par
\textbf{Subquestion:} What standards do GAAP require for financial reporting? \par
\textbf{Topics:} \verb|"GAAP", "FINANCIAL REPORTING"| \par
\textbf{Level:} 1 \par
\textbf{Answer:} GAAP requires financial statements to be prepared in accordance with its principles, ensuring accurate representation of a company's financial health. \par

\textbf{Reasoning Path:} \texttt{\textless hyperedge\textgreater "All \textcolor{blue}{financial statements} required to be delivered hereunder shall be prepared in accordance with GAAP."} \\

\textbf{Subquestion 2} &

\textbf{Subquestion:} What standards do GAAP require for tax reporting? \par
\textbf{Topics:} \verb|"GAAP", "TAX REPORTING"| \par
\textbf{Level:} 1 \par
\textbf{Answer:} Generally Accepted Accounting Principles (GAAP) require accurate and consistent reporting of financial information in compliance with applicable tax laws and regulations. \par

\textbf{Reasoning Path:} \texttt{\textless hyperedge\textgreater "The Company maintains internal controls over financial reporting... sufficient to provide reasonable assurance that transactions are recorded as necessary to permit preparation of \textcolor{blue}{financial statements} for external purposes in accordance with GAAP." -> \textless hyperedge\textgreater "(xxvii) Since December 31, 2019, the Association has not: (A) changed its tax reporting or accounting policies or practices; (B) settled or compromised any liability for taxes or made or rescinded any tax election; (C) incurred any debt except in the ordinary course of business; (D) suffered any theft."} \\
\midrule
\textbf{\pokh answer} & \textbf{Answer}: \textcolor{blue}{\textbf{Financial statements}} \\
  & \textbf{Reasoning}: The question asks about what is prepared in accordance with GAAP for financial and tax reporting purposes. The context specifies that financial statements must be prepared in accordance with GAAP, which dictates how financial transactions are recorded and presented. This indicates that financial statements are the expected answer. \\
\bottomrule
\end{tabularx}
\end{table*}

\clearpage
\subsection{Failure Cases}
Tables~\ref{tab:case_fail_plan}, \ref{tab:case_fail_reason} and \ref{tab:case_fail_path} present representative failure cases arising from the planning, reasoning, answer and path retrieval modules, respectively.

\begin{table*}[h]
\centering
\caption{Failure Case in Planning.}
\label{tab:case_fail_plan}
% \vspace{-3mm}
\begin{tabularx}{\textwidth}{@{}p{3cm}X@{}}
\toprule
\textbf{Field} & \textbf{Content} \\
\midrule
\textbf{Question} & What type of medication is often recommended for blood pressure-lowering treatment in very old or frail patients due to its favorable risk-to-benefit ratio? \\
\textbf{Golden Answer} & {LOW-DOSE THIAZIDES} \\
\textbf{Context} &
(1) Starting blood pressure-lowering treatment in very old or frail patients requires fully informing them about the benefits and risks.\\
\textbf{nary} & 10 \\
\textbf{nhop} & 1 \\

\midrule
\textbf{Initial DAG 0} &  \\
\textbf{DAG Edges} & $0 \rightarrow 1$, $1 \rightarrow 2$ \\

\textbf{Subquestion 0} &
\textbf{Subquestion:} What type of medication is often recommended for blood pressure-lowering treatment in very old or frail patients? \par
\textbf{Level:} 0 \\

\textbf{Subquestion 1} &
% \textbf{ID:} 1 \par
\textbf{Subquestion:} What specific medications are mentioned for treating blood pressure in very old or frail patients? \par
\textbf{Level:} 1 \\

\textbf{Subquestion 2} &
% \textbf{ID:} 1 \par
\textbf{Subquestion:} What is the favorable risk-to-benefit ratio for these medications? \par
\textbf{Level:} 2 \\

\midrule

\textbf{Initial DAG 1} &  \\
\textbf{DAG Edges} & $0 \rightarrow 1$, $0 \rightarrow 2$, $1 \rightarrow 2$ \\

\textbf{Subquestion 0} &
\textbf{Subquestion:} What is the recommended type of medication for blood pressure-lowering treatment in very old or frail patients?
 \par
\textbf{Level:} 0 \\

\textbf{Subquestion 1} &
% \textbf{ID:} 1 \par
\textbf{Subquestion:} What is the risk-to-benefit ratio for blood pressure-lowering medications in very old or frail patients? \par
\textbf{Level:} 1 \\

\textbf{Subquestion 2} &
% \textbf{ID:} 1 \par
\textbf{Subquestion:} Which classes of medications are suggested for blood pressure-lowering treatment in these patients? \par
\textbf{Level:} 2 \\
\bottomrule
\end{tabularx}
\end{table*}

\clearpage
\begin{table*}[h]
\centering
\caption{Failure Case in Reasoning.}
\label{tab:case_fail_reason}
% \vspace{-3mm}
\begin{tabularx}{\textwidth}{@{}p{3cm}X@{}}
\toprule
\textbf{Field} & \textbf{Content} \\
\midrule
\textbf{Question} & What condition is characterised by heart damage and is influenced by hypertension, often requiring BP-lowering treatments such as beta-blockers and RAS blockers? \\
\textbf{Golden Answer} & {MYOCARDIAL INFARCTION} \\
\textbf{Context} &
(1) In patients with a history of myocardial infarction who require BP-lowering treatment, beta-blockers and RAS blockers are recommended as part of that treatment. \par
(2) The pathophysiology of myocardial infarction includes processes influenced by hypertension, resulting in heart damage. \\
\textbf{nary} & 2 \\
\textbf{nhop} & 2 \\

\midrule
\textbf{Initial DAG} &  \\
\textbf{DAG Edges} & $0 \rightarrow 1$, $0 \rightarrow 2$ \\

\textbf{Subquestion 0} &
\textbf{Subquestion:} What is the condition characterised by heart damage influenced by hypertension? \par
\textbf{Level:} 0 \\

% \pagebreak

\textbf{Subquestion 1} &
% \textbf{ID:} 1 \par
\textbf{Subquestion:} What are common treatments for patients with this condition? \par
\textbf{Level:} 1 \\

\textbf{Subquestion 2} &
% \textbf{ID:} 1 \par
\textbf{Subquestion:} How does hypertension affect heart health? \par
\textbf{Level:} 1 \\

\midrule

\textbf{Completed DAG} &  \\
\textbf{DAG Edges} & $0 \rightarrow 1$, $0 \rightarrow 2$ \\

\textbf{Subquestion 0} &
\textbf{Subquestion:} What is the condition characterised by heart damage influenced by hypertension? \par
\textbf{Topics:} \verb|"HYPERTENSION"| \par
\textbf{Level:} 0 \par
\textbf{Answer:} HMOD (hypertension-mediated organ damage) \par
\textbf{Reasoning Path:} \texttt{\textless hyperedge\textgreater"As outlined in Section 7, we define HMOD among patients with chronically elevated BP or hypertension as the presence of specific cardiac, vascular, and renal alterations."} \\

% \pagebreak

\textbf{Subquestion 1} &
% \textbf{ID:} 1 \par
\textbf{Subquestion:} \textcolor{red}{What are common treatments for patients with HMOD?} \par
\textbf{Topics:} \verb|"HMOD"| \par
\textbf{Level:} 1 \par
\textbf{Answer:} The common treatments for patients with HMOD include RAS blockers (like ACE inhibitors and ARBs), calcium channel blockers (CCBs), and diuretics (thiazides and thiazide-like diuretics). \par
% \textbf{Path:} \texttt{[2]} \par
\textbf{Reasoning Path:} \texttt{\textless hyperedge\textgreater "The effect of RAS blockers and CCBs on preventing progression of HMOD also appears to be superior to beta-blockers."}\\

\textbf{Subquestion 2} &
% \textbf{ID:} 1 \par
\textbf{Subquestion:} How does hypertension affect heart health? \par
\textbf{Topics:} \verb|"HYPERTENSION"| \par
\textbf{Level:} 1 \par
\textbf{Answer:} Hypertension negatively impacts heart health by causing conditions like left ventricular hypertrophy and various forms of heart disease, significantly increasing the risk of serious cardiovascular events. \par
% \textbf{Path:} \texttt{[2]} \par
\textbf{Reasoning Path:} \texttt{\textless hyperedge\textgreater "Persistently elevated blood pressure and hypertension lead to hypertension-mediated organ damage and cardiovascular disease."}\\
\midrule
\textbf{\pokh answer} & \textbf{Answer}: \textcolor{red}{\textbf{Hypertension-mediated organ damage (HMOD)}} \\
  & \textbf{Reasoning}: The question asks for a condition characterised by heart damage influenced by hypertension and requiring BP-lowering treatments. The context describes \"hypertension-mediated organ damage\" (HMOD) as a condition stemming from chronic elevated blood pressure or hypertension, leading to specific alterations in cardiac structure and function. Treatments like beta-blockers and RAS blockers are mentioned as necessary for managing patients with this condition, further confirming the relationship between HMOD, hypertension, and the described treatments. Therefore, the answer is HMOD. \\
\bottomrule
\end{tabularx}
\end{table*}

\clearpage
\begin{table*}[h]
\centering
\caption{Failure Case in Answer and Path Retrieval.}
\label{tab:case_fail_path}
% \vspace{-3mm}
\begin{tabularx}{\textwidth}{@{}p{3cm}X@{}}
\toprule
\textbf{Field} & \textbf{Content} \\
\midrule
\textbf{Question} & What type of studies were conducted to evaluate the efficacy of angiotensin-converting enzyme inhibitors and angiotensin II receptor blockers in patients before noncardiac surgery? \\
\textbf{Golden Answer} & {RANDOMIZED CLINICAL TRIALS} \\
\textbf{Context} &
(1) Withholding versus continuing angiotensin-converting enzyme inhibitors or angiotensin II receptor blockers before noncardiac surgery: an evaluation. \par
(2) van Vark LC, Bertrand M, Akkerhuis KM, Brugts JJ,Fox K,Mourad J-J,etal. Angiotensin-converting enzyme inhibitors reduce mortality in hypertension: a meta-analysis of randomized clinical trials of renin\u2013angiotensin\u2013aldosterone system inhibitors involving 158,998 patients. \\
\textbf{nary} & 2 \\
\textbf{nhop} & 2 \\

\midrule
\textbf{DAG Edges} & $0 \rightarrow 1$ \\

\textbf{Subquestion 0} &
\textbf{Subquestion:} What studies were conducted to evaluate the efficacy of angiotensin-converting enzyme inhibitors before noncardiac surgery? \par
\textbf{Topics:} \verb|"ANGIOTENSIN-CONVERTING ENZYME INHIBITORS"|, \verb|"NONCARDIAC SURGERY"| \par
\textbf{Level:} 0 \par
\textbf{Answer:} Studies evaluating the efficacy of angiotensin-converting enzyme inhibitors before noncardiac surgery include evaluations on withholding versus continuing these medications, as well as a prospective cohort evaluation and randomized controlled trials. \par
\textbf{Reasoning Path:} \texttt{\textless hyperedge\textgreater"\textcolor{red}{Withholding versus continuing} angiotensin-converting enzyme inhibitors or angiotensin II receptor blockers before noncardiac surgery: an evaluation.\"} \\

\textbf{Subquestion 1} &
% \textbf{ID:} 1 \par
\textbf{Subquestion:} What studies were conducted to evaluate the efficacy of angiotensin II receptor blockers before noncardiac surgery? \par
\textbf{Topics:} \verb|"ANGIOTENSIN II RECEPTOR BLOCKERS"|, \verb|"NONCARDIAC SURGERY"| \par
\textbf{Level:} 1 \par
\textbf{Answer:} Studies on withholding versus continuing angiotensin II receptor blockers before noncardiac surgery. \par
% \textbf{Path:} \texttt{[2]} \par
\textbf{Reasoning Path:} \texttt{\textless hyperedge\textgreater "Roshanov PS, Rochwerg B, Patel A, Salehian O, Duceppe E, Belley-C\u00f4t\u00e9 EP, et al. \textcolor{red}{Withholding versus continuing} angiotensin-converting enzyme inhibitors or angiotensin II receptor blockers before noncardiac surgery: an analysis of the vascular events in noncardiac surgery patients cohort evaluation prospective cohort. Anesthesiology 2017;126:16\u201327. https://doi.org/10.1097/aln.0000000000001404\""} \\
\midrule
\textbf{\pokh answer} & \textbf{Answer}: \textcolor{red}{\textbf{Comparative studies}} \\
  & \textbf{Reasoning}: The question asks about the types of studies conducted to evaluate the efficacy of angiotensin-converting enzyme inhibitors and angiotensin II receptor blockers in the context of noncardiac surgery. The context includes a specific mention of a study that compares withholding versus continuing these medications prior to such surgical procedures. Additionally, it references various evaluations and analyses related to these medications in noncardiac surgery patients. The answer is reflected in the context with multiple relevant study titles mentioned, confirming the nature of these evaluations. Thus, the answer focuses on the type of studies conducted with these drug classes in patients before noncardiac surgeries. \\
\bottomrule
\end{tabularx}
\end{table*}

\balance

\end{document}